\documentclass[manuscript]{acmart}
\AtBeginDocument{%
  \providecommand\BibTeX{{%
    \normalfont B\kern-0.5em{\scshape i\kern-0.25em b}\kern-0.8em\TeX}}}

\usepackage{graphicx}
\usepackage{xcolor}
\usepackage{hyperref}
\usepackage{textcomp} 
\usepackage{float}
\usepackage{wrapfig}
\usepackage{amsmath, amsthm, amsfonts}

\usepackage{longtable}

\newcommand{\dd}{\scriptsize{\textnormal{d}}}

\newcommand{\Glp}{\textnormal{GL}^+}
\newcommand{\Symp}{\textnormal{Sym}^+}

\newcommand{\SO}{\textnormal{SO}}

\definecolor{beige}{RGB}{153, 143, 110}
\definecolor{midnightblue}{RGB}{25,25,112}
\definecolor{navyblue}{RGB}{0,0,128}

\DeclareRobustCommand{\ShowColormap}{\raisebox{-0.14em}{\includegraphics[height=.8em]{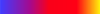}}}

\theoremstyle{definition}
\newtheorem{exmp}{Example}[section]

\setcopyright{acmcopyright}
\copyrightyear{2023}
\acmYear{2023}
\acmDOI{XXXXXXX.XXXXXXX}

\acmJournal{JOCCH}

\begin{document}

\title{Intrinsic shape analysis in archaeology: A case study on ancient sundials}

\author{Martin Hanik}
\authornote{Corresponding author.}
\email{hanik@zib.de}
\orcid{0000-0002-7120-4081}
\affiliation{%
  \institution{Freie Universit\"at Berlin}
  \streetaddress{Kaiserswerther Str. 16--18}
  \city{Berlin}
  \country{Germany}
  \postcode{14195}
}

\author{Benjamin Ducke}
\email{benjamin.ducke@dainst.de}
\orcid{0000-0002-7120-4081}
\affiliation{%
  \institution{German Archaeological Institute}
  \streetaddress{Podbielskiallee 69--71}
  \city{Berlin}
  \country{Germany}
  \postcode{14195}
}

\author{Hans-Christian Hege}
\email{hege@zib.de}
\orcid{0000-0002-6574-0988}
\affiliation{%
  \institution{Zuse Institute Berlin}
  \streetaddress{Takustr. 7}
  \city{Berlin}
  \country{Germany}
  \postcode{14195}
}

\author{Friederike Fless}
\email{praesidentin@dainst.de}
\orcid{0000-0003-4500-7778}
\affiliation{%
  \institution{German Archaeological Institute}
  \streetaddress{Podbielskiallee 69--71}
  \city{Berlin}
  \country{Germany}
  \postcode{14195}
}

\author{Christoph von Tycowicz}
\email{tycowicz@zib.de}
\orcid{0000-0002-1447-4069}
\affiliation{%
  \institution{Freie Universit\"at Berlin}
  \streetaddress{Kaiserswerther Str. 16--18}
  \city{Berlin}
  \country{Germany}
  \postcode{14195}
}

\renewcommand{\shortauthors}{Hanik et al.}

\begin{abstract}
\textcolor{black}{The fact that the physical shapes of man-made objects are subject to overlapping influences---such as technological, economic, geographic, and stylistic progressions---holds great information potential.
On the other hand, it is also a major analytical challenge to uncover these overlapping trends and to disentagle them in an \emph{unbiased} way.}
This paper explores a novel mathematical approach to extract archaeological insights from \textcolor{black}{ensembles of similar} artifact shapes.
We show that by considering all shape information in a \textit{find collection}, it is possible to identify shape patterns that would be difficult to discern \textcolor{black}{by considering the artifacts individually or} by classifying shapes into predefined archaeological types and analyzing the associated distinguishing characteristics.	

Recently, series of high-resolution digital representations of artifacts have become available. Such data sets enable the application of extremely sensitive and flexible \textcolor{black}{methods of shape} analysis. We explore this \textcolor{black}{potential} on a set of 3D models of ancient Greek and Roman sundials, with the aim of providing alternatives to the traditional archaeological method of ``trend extraction by ordination'' (typology).\\
In \textcolor{black}{the proposed} approach, each 3D shape \textcolor{black}{is represented} as a point in a \emph{shape space}\textcolor{black}{---}a high-dimensional, curved, non-Euclidean space. Proper consideration of its mathematical properties reduces bias in data analysis and thus improves analytical power. 
By performing regression in shape space, we find that for Roman sundials, the bend of the shadow-receiving surface of the sundials changes with the latitude of the location. This suggests that, apart from the inscribed hour lines, also a sundial's shape was adjusted to the place of installation.
As an example of more advanced inference, we use the identified trend to infer the latitude at which a sundial, whose location of installation is unknown, was placed.

We also derive a novel method for differentiated morphological trend assertion, building upon and extending the theory of geometric statistics and shape analysis. Specifically, we present a regression-based method for statistical normalization of shapes that serves as a means of disentangling \textit{parameter-dependent effects (trends)} and \textit{unexplained variability}. 
In addition, we show that this approach is robust to noise in the digital reconstructions of the artifact shapes.
\end{abstract}



\maketitle

\section{Introduction}

Within the discipline of archaeology, artifacts, i.e., the physical remains of tools, weapons, clothing, adornments and other man-made objects, represent the largest and most diverse source of evidence. Notwithstanding the importance of ornamentation, use-wear traces, material composition, etc.\, the aspect of shape has long served as the primary means of establishing structure in both time (\textit{chronology}) and space (\textit{chorology}) within the vast universe of known artifacts \cite{Montelius1903}. In this section, we introduce the basic premises of a novel mathematical approach to unravel the manifold sources of information contained in the shapes of artifacts.

\subsection{Shape and archaeology}

Indeed, shape has always been fundamental to the analysis of archaeological artifacts. While this can also be said about the objects of interest of related disciplines, such as paleontology, geology and biology, the ``human dimension'' of artifact creation adds complex layers of technological, economic, artistic and other social components to the physical manifestations of shapes. As a consequence, archaeology---a discipline that has its roots equally in the prospering natural sciences and antiquarian movement of the 18th century~\cite[Ch. 1]{Renfrew2019}---traditionally uses the method of \textit{typology} to impose a more or less well-defined order, called \textit{typological sequence}~\cite[Ch. 4]{Renfrew2019}, on a series of artifacts. It is hypothesized to isolate the dominant morphological trends and provide a basis for extraction and further, informal analysis of the unexplained residual trends: e.g., geographic detrending through spatial ordering facilitates the detection of chronological trends; additional temporal ordering may reveal social stratification. There is a large body of published research on the process of typological ordering (often called \textit{seriation} in archaeology and \textit{ordination} in ecology~\cite{OBrien1998}), including computational methods of varying complexity (see, e.g., \textcolor{black}{Refs.}~\cite{Laxton1989,Madsen1989,Scott1992}). The combinatorial complexity of seriation implies that optimal results are computationally infeasible even for a relatively small number of artifacts, whereas archaeological excavations often yield tens of thousands of artifacts. Consequently, no single procedure is undisputed and manual grouping and ordering based on subjective mixtures of features (artistic/stylistic, technological, functional, etc.) is still common, as are simplified and abstract graphical representations (drawings) of objects. These discretizations, of both concepts and representations, into archaeological ``types'', i.e., idealized forms that serve as representatives of classes of objects in a particular spatio-temporal context, are not free of arbitrariness, and they imply a loss of information.

The resulting limitations are becoming more apparent as new image-based technologies provide economical means for digitally capturing and representing shapes with full 3D detail and high accuracy. 
Our contribution aims to use innovative mathematical approaches to enable the transition from type-based analysis of archaeological object shapes to a geometric, purely data-driven analysis. \textcolor{black}{We focus on knowledge inference from a whole \textit{set} of artifact shapes in order to demonstrate the potential of modern digital repositories.}
We use the example of ancient sundial surfaces from Italy and Greece. These are well suited to our approach because their physical shape must necessarily reflect at least one controllable trend, that of the (intended) geographic location of use, in order to function properly.
We ignore \textit{a priori} typological categories and base our analysis solely on the shapes themselves and their derived geometric properties. We do, however, make use of geographical (object find spots) and, as part of an application example, cultural (Greek, Roman) contextual information provided along with the 3D data, to drive our analysis. The 3D models of the sundials were generated by an image-based 3D reconstruction process called ``Structure from Motion'' and ``Multi-view Stereo'' (SfM/MVS), followed by surface reconstruction and optional texturing to enhance visual detail, but without effecting geometric properties. This process will not be discussed in any detail here, as it is well known also in archaeology (see, e.g., \textcolor{black}{Refs.}~\cite{Green2018,Ducke2018}). 
It is important to note that the quality of the resulting data\footnote{The 3D data sets of sundials used have been published as open data by the Topoi research cluster (see App.~\ref{app:ID}).} is more than sufficient for the presented approach, and that the actual scale of the objects does not matter.

\subsection{Shape and mathematical shape space}

The starting point for our approach is a shape space: Although each artifact is embedded in 3D Euclidean space (with coordinates resulting from physical data acquisition), the set of all its surface boundary points can be considered as a single point in a high-dimensional, curved space; this is the \emph{shape space}\footnote{More precisely, there are several (known) ways to encode the surface boundaries as points in generally different high dimensional, curved spaces, which are all called shape spaces; which one to choose depends on the application at hand.}. For an introduction see, e.g., the textbooks~\cite{DrydenMardia2016, Kendall_ea2009}. The shape space is key to understanding fundamental advantages of our approach over more traditional statistical methods.

The basic challenge is to separate the influence of variables of interest on the shapes of the artifacts from noise (which includes data acquisition and modeling errors as well as the influence of unexplained or irrelevant variance). This task, of course, falls into the classic domain of multivariate statistics, which provides a wealth of methods for identifying key trends in data as well as ordering and grouping them according to robust criteria. At its core, the highly developed multivariate apparatus is based on the analysis of the variance of the input data---measured in Euclidean distances. Methods such as principal components analysis, clustering analysis or correspondence analysis (a method that has gained much popularity in archaeology~\cite{Greenacre1984,Cool1995}) all address the problem of sorting data into groups (alternatively: along components, factors, etc.) with minimal internal (``in group'') and maximal external (``between groups'') difference (``variance''). 
While the assumption of ``Euclideanness" simplifies the underlying mathematics and provides robust results, there is increasing evidence that it limits the ability to capture variations in shapes~\cite{Kendall_ea2009,BauerBruverisMichor2014,vonTycowicz_ea2018}.

But even if the assumptions of Euclideaness and global orthogonality are relaxed, the fundamental issue remains that the resulting groups (components, factors, etc.) represent synthetic variables, determined by purely formal criteria, that have no immediately meaningful relations with any trends that were identified \textit{a priori}.
\textcolor{black}{Using conventional multivariate statistics,} it is very hard to isolate and identify subtle contributions (e.g., by various geographic, social or technological processes) in such complex physical manifestations as the shapes of artifacts or architectural elements. These problems can be alleviated by analyzing shapes directly, without grossly \textcolor{black}{simplistic} transformations and the associated loss of information. 
When applied to geometric objects, this requires a notion of space that provides a complete structural characterization of their shape along with an appropriate measure of (dis)similarity that allows subtle and individual trends \textcolor{black}{to be assessed and isolated.}
Although such a treatment leads to high-dimensional and nonlinear problems, efficient implementations can be obtained employing a recent \textcolor{black}{type of} shape space~\cite{vonTycowicz_ea2018} that provides closed-form expressions for intrinsic operations and, thus, simple and fast algorithms.
To give the interested reader an easy way to test the proposed methods and to ensure the reproducibility of our results, we provide a Python implementation as part of the Morphomatics library~\cite{Morphomatics}; notebooks with our experiments can be found online under \url{https://github.com/morphomatics/SundialAnalysis}. 

By utilizing curved shape spaces, we leave multivariate statistics in its traditional form.
Remarkably, employing differential geometric concepts, we can obtain generalized statistical tools \cite{GrohsHollerWeinmann2020, Pennec_ea2019_book} that take the rich geometric structure of shape space into account (i.e.,\ dispense with linearization), while being compatible with the Euclidean counterparts (i.e.,\ agree with the original definition for the special case of flat vector spaces).
The transition to curved spaces with the analytical methods available there allows for more flexible derivations and higher analytical sensitivity than multivariate statistics with its dependence on global \textcolor{black}{Euclideaness}~\cite{AmbellanZachowvonTycowicz2019,Huckemann_ea2010,vonTycowicz_ea2018}.

\begin{figure*}[ht]
    \centering
    \includegraphics[width=1\textwidth]{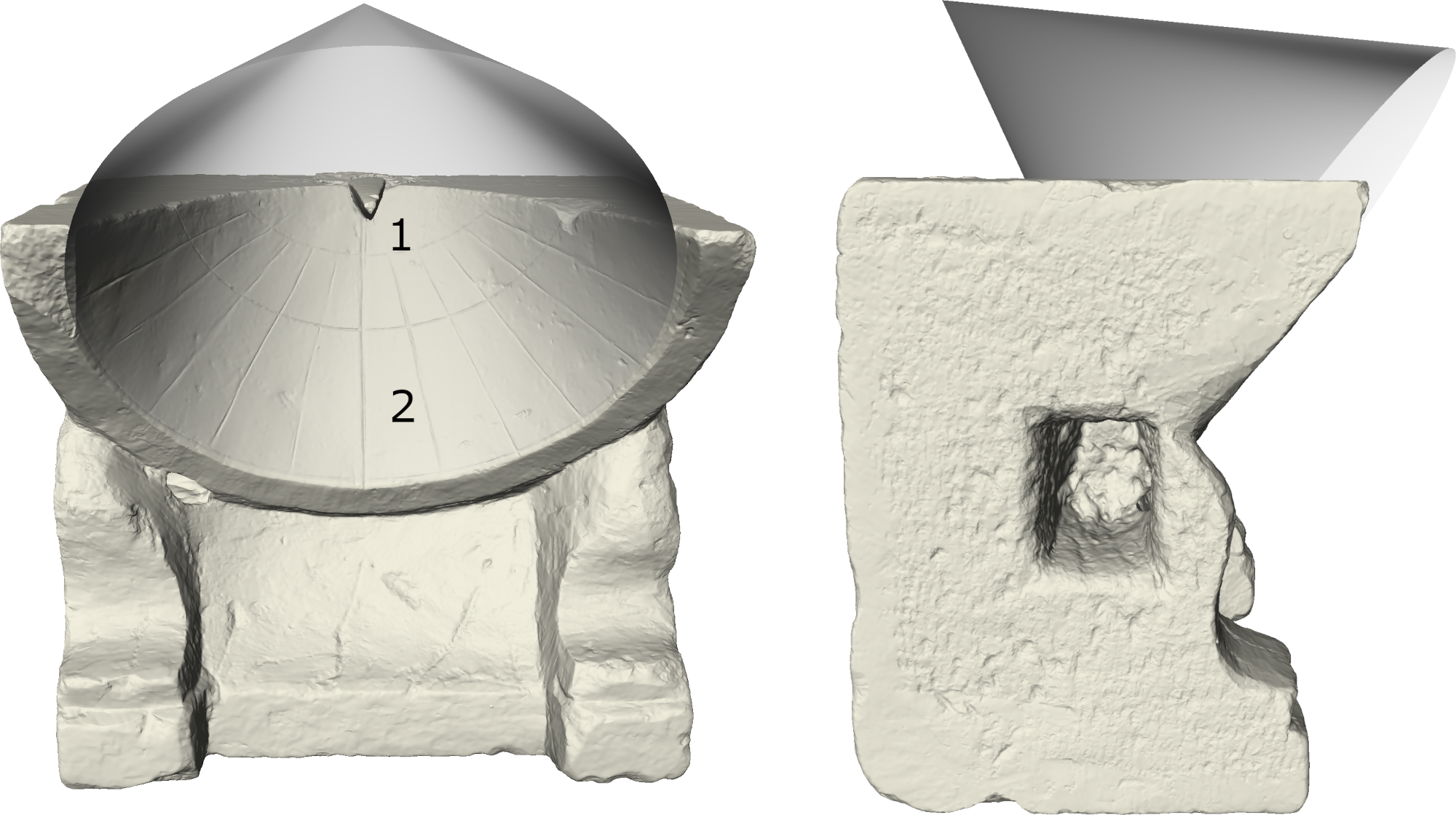}
    \caption{Front (left) and side (right) view of a conical sundial with gnomon (1) and shadow surface (2). \textcolor{black}{In order to craft the shadow surface, part of a cone was cut from the stone, as indicated by the gray cone in the picture.}}\label{fig:conical_sundial}
    \Description[View of conical sundial]{Front and side view of conical sundial gnomon}
\end{figure*}
\begin{figure*}[ht]
    \centering
    \includegraphics[width=1\textwidth]{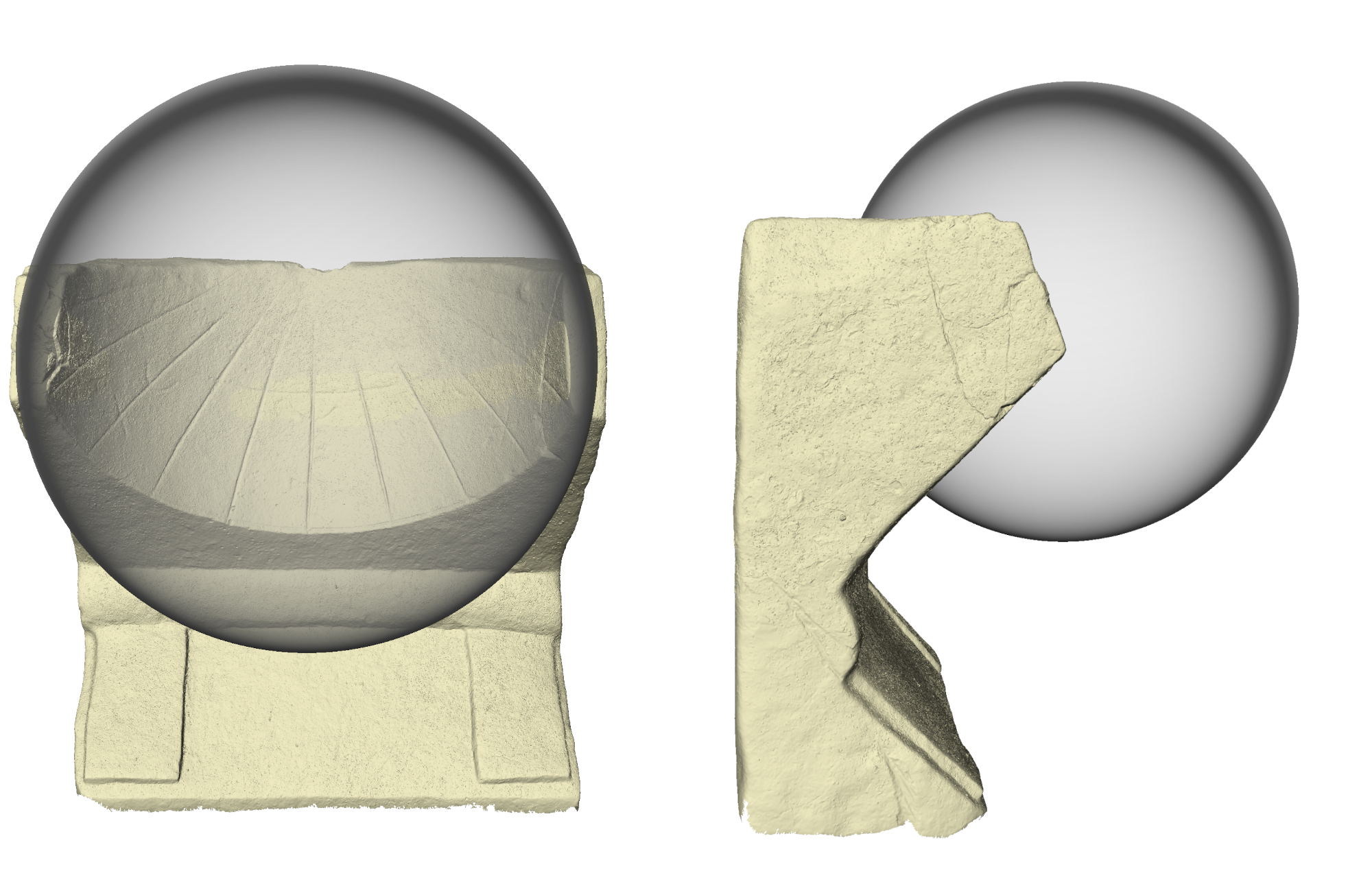}
    \caption{Front (left) and side (right) view of a spherical sundial with missing gnomon. \textcolor{black}{In order to craft the shadow surface, part of a round ball was cut from the stone, as indicated by the gray sphere.}}\label{fig:sperical_sundial}
    \centering
    \Description[View of spherical sundial]{Front and side view of spherical sundial without gnomon}
\end{figure*}

\subsection{Problem, analysis task} \label{sec:problem}

In antiquity, sundials were mainly used for the measurement of time~\cite{Schaldach2016}. While the sun is above the horizon, sundials indicate the time by the position of the shadow of a so-called \textit{gnomon} on a shadow-receiving surface; see Fig.~\ref{fig:conical_sundial}. In this article, we adopt the term \textit{shadow surface} for the latter.  
Shape-wise, at least four types of sundials can be distinguished~\cite{RinnerFritschGrasshoff2013}: conical, spherical, cylindrical and planar sundials---determined solely by the shape of the shadow surface.
In Figs.~\ref{fig:conical_sundial} and \ref{fig:sperical_sundial}, we display a conical and a spherical sundial. Independently from the type, the geographical latitude had to be considered during construction to ensure proper \textcolor{black}{functioning}. In particular, it is well known that the \emph{hour lines}, which were inscribed into the shadow surface to read off time, were adapted to the latitude of the location of installation~\cite{Jones2019,Huettig2000}. 
An obvious question then is whether other parts have also been changed: By identifying a latitude-dependent trend in the data, we show that, at least for the spherical type, it is very likely that the \textit{shape} of the shadow surfaces was also adapted by the craftsman. We demonstrate an application of this trend by inferring the latitude of the place of installation of a sundial with uncertain geographical location of use.
Furthermore, if one wants to compare shapes of shadow surfaces from different geographical locations, e.g., with respect to differences in construction principles, the shapes should be \emph{normalized} to latitude; otherwise, one might misinterpret differences that actually are ``imposed'' on the craftsman by the latitude of the intended place of use. In this work we develop the necessary methodology to perform this normalization. We then use it for an illustrative analysis, comparing the mean shapes of shadow surfaces from ancient Greek and Roman sundials. With more (reliable) data, such an analysis could---unbiased from latitudinal effects---reveal differences in construction principles between the groups.


\noindent
To sum up, our contributions are the following. We
\begin{itemize}
    \item demonstrate the use of statistical tools for shape data in archaeology,
    \item identify a latitude-driven shape trend for the shadow surface of ancient Roman sundials, 
    \item use the trend to approximate the latitude of the place of installation of a sundial for which the latter is uncertain,
    \item present a new method for the analysis of parameter-dependent shape data that allows separating different influencing variables such as time and location.
\end{itemize}

The article is structured as follows: In Sec.~\ref{sec:method}, we describe the method. This includes the data and the pre-processing we performed as well as the shape space and the mathematical methods that we used. We then discuss the results in Sec.~\ref{sec:results}
and give a short conclusion in Sec.~\ref{sec:conclusion}.


\section{Method} \label{sec:method}

In this section, we describe the \textcolor{black}{analysis} pipeline that we applied to analyze the shapes of the shadow surfaces. After presenting the data and the performed preprocessing, we describe the mathematical shape representation that was used, and discuss statistical methods (in particular, generalizations of the mean and linear regression) that allow us to study sets of shapes. Furthermore, we introduce a novel method for normalizing shapes with respect to a given parameter and exemplify its use for studying residual shape variations.
Since the normalization has not been published before, we give the mathematical details; further mathematical background is detailed in the Appendix.

\subsection{Data and data preprocessing} \label{sec:data}

Today, about 500 ancient Greek or Roman sundials are known and 3D models of many of them can be found in the repository of the Topoi Excellence of Cluster~\cite{Grasshof_ea2016_data}. They were reconstructed in a Structure from Motion/Multiview Stereo (SfM/MVS) procedure\textcolor{black}{, as} detailed in \textcolor{black}{Ref.}~\cite{FritschRinnerGrasshoff2013}. For this study, we used spherical sundials from Greece and the Italian peninsula and considered all available models with well preserved shadow surface; their IDs and sites are given in App.~\ref{app:ID}. 

\begin{figure*}[ht]
\centering
    \includegraphics[width=.8\textwidth]{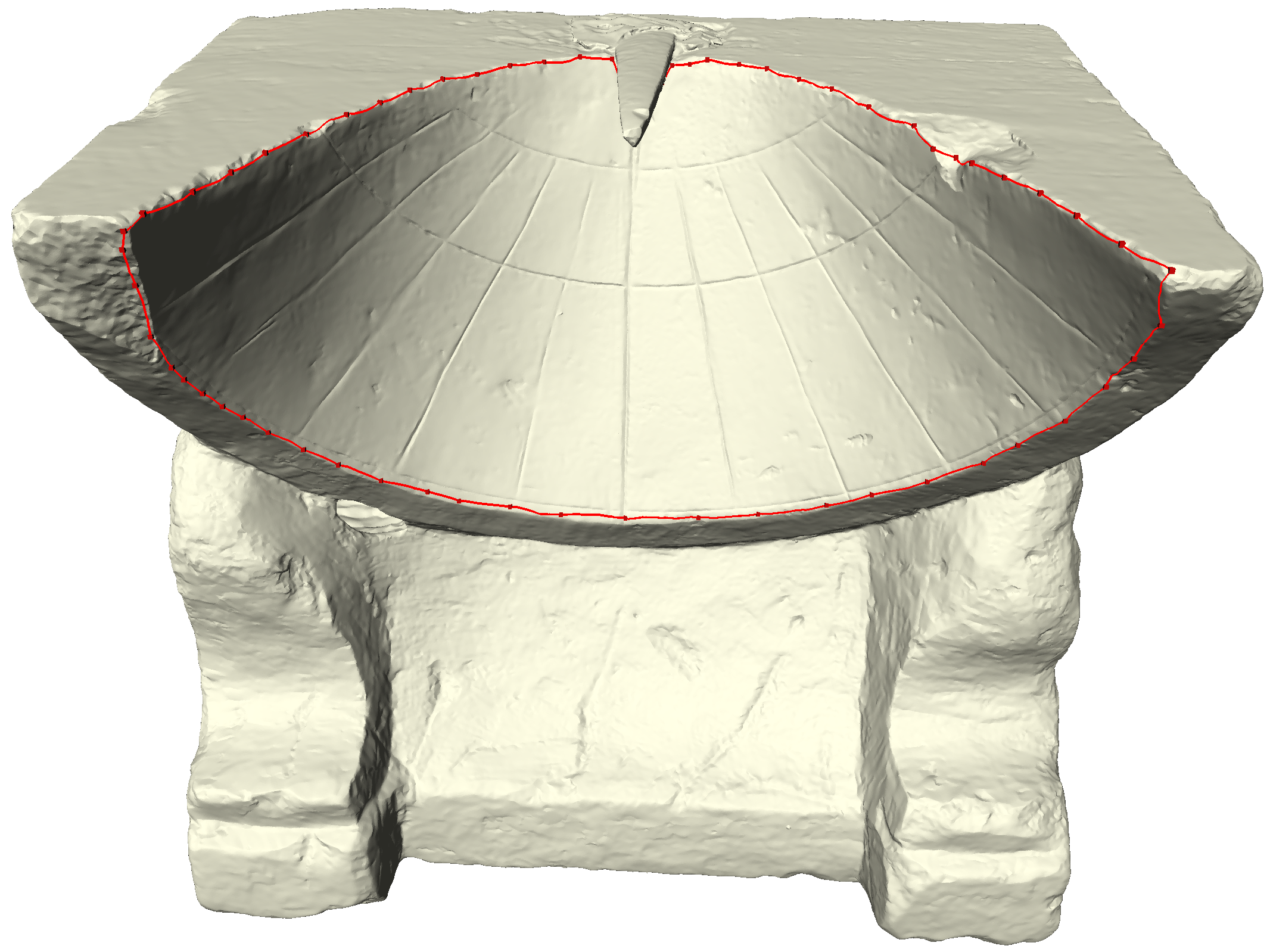}
\caption{Manually segmented shadow surface}\label{fig:pathset}
\Description[Segmented shadow surface]{Spherical sundial with manually segmented shadow surface}
\hfill
    \includegraphics[width=1\textwidth]{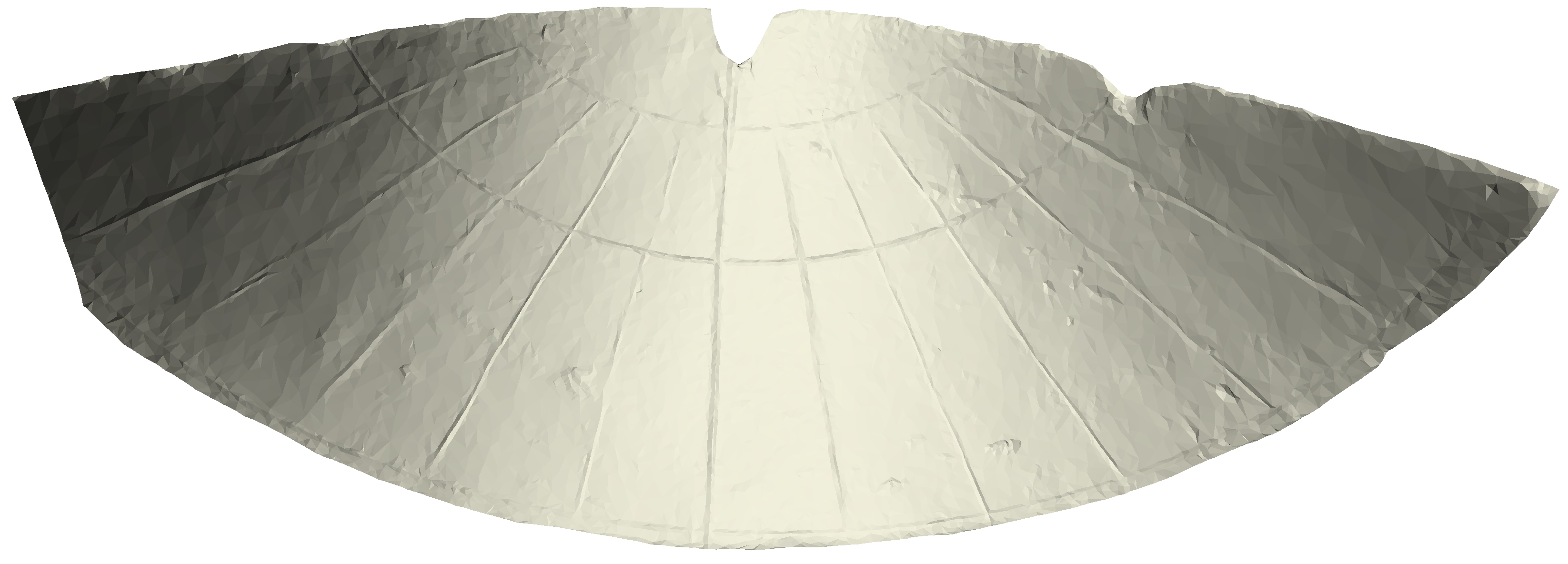}
\caption{Extracted shadow surface}\label{fig:shadow_surface}
\Description[Extracted shadow surface]{Extracted shadow surface with regularized boundary triangles}
\end{figure*}

As for data preparation, we extracted the shadow surfaces with the software Amira~\cite{Stalling2005}.
We manually selected the shadow surfaces and corrected triangles near the new boundary that had extreme angles.
Furthermore, the resolution of the surface meshes was reduced to about $20\,k$ faces using the \textit{quadric edge collapse decimation scheme} implemented in the free software MeshLab~\cite{Cignoni2008}.
An example of a segmented shadow surface can be seen in Fig.~\ref{fig:pathset}; it is shown after extraction and simplification in Fig.~\ref{fig:shadow_surface}.

To perform statistics on shapes we rely on a group-wise correspondence, i.e., consistent point-to-point relationships of the meshes. \textcolor{black}{There are many approaches to establish such correspondences, e.g., non-rigid registration~\cite{BronsteinBronsteinKimmel2008,Tam_ea2012} and functional map--based~\cite{Ovsjanikov_ea2016} approaches.
We obtained one by choosing a shadow surface as reference and registering it with a harmonic map to all other surfaces through the following two steps:}
First, we registered the boundaries by using 3 landmarks.
While the two geodesically farthest points were automatically approximated as the global minimizer and maximizer of the surface's Fiedler vector (the eigenvector corresponding to the second smallest eigenvalue of the surface's Laplace-Beltrami operator), the third landmark was manually selected as the lowest point of the gnomon hole.
(The gnomon itself was missing at all but one of the sundials under study.)
Then, we established the correspondence of the interior part using the \textit{discrete harmonic map for disc-like topology} as described in \textcolor{black}{Ref.}~\cite{brett2000construction}. Finally, effects of rotation and scale were removed by a group-wise Procrustes alignment of the meshes~\cite{Goodall1991}.
The processed meshes are available online; see~\cite{HanikvonTycowicz2022}.

Note that on many sundials the upper left and right corners are only poorly preserved, showing a varying degree of decay.
Thus, the correspondences of the worn edges (which are rather small in comparison to the rest of the sundial) are only approximate. In the later analysis we therefore concentrated on the other areas (that are conserved very well).

\subsection{Shape spaces} \label{sec:shape_space}
Informally in mathematics, a ``shape'' is the set of all the geometric properties of an object that are invariant under similarity transformations, i.e., under translations, rotations and scalings. 
\textcolor{black}{To analyze sets of shapes one needs the notion of shape spaces, i.e., spaces in which shapes naturally ``live'', that is, in which each point represents a full geometric shape. (For a good references on the analysis of individual shapes see \textcolor{black}{Ref.}~\cite{Biasotti_ea2014}.)} There are different approaches to construct such shape spaces and to utilize them, depending on application characteristics. A classical example is Kendall shape space~\cite{Kendall_ea2009,DrydenMardia2016}, which relies on landmark configurations. 
An overview of shape spaces whose elements are diffeomorphisms (i.e., differentiable, bijective maps with differentiable inverses) can be found in~\cite{BauerBruverisMichor2014}---with more in-depth discussions, e.g., in \textcolor{black}{Refs.}~ \cite{SrivastavaKlassen2016,Younes2019}. Skeletal models are discussed in \textcolor{black}{Refs.}~\cite{SiddiqiPizer2008,Pizer_ea2020}.
Finally, physically-based spaces were investigated in \textcolor{black}{Refs.}~\cite{heeren2014exploring, vonTycowicz_ea2018, AmbellanZachowvonTycowicz2019}. Almost all of the above shape spaces have an inherent non-Euclidean structure. Luckily, techniques from the mathematical field of Riemannian geometry (see App.~\ref{app:math_detail} for an introduction) can either be directly applied (since shape spaces are so-called manifolds) or have been successfully extended to them. This fact is utilized by geometric statistics (see Sec.~\ref{sec:statistics}) to generalize concepts from multivariate statistics for the analysis of shapes. 

Before proceeding, we would like to emphasize that the methods we present in this article are not limited to any particular choice of shape representation; we require only that the shapes are modeled as elements of a Riemannian manifold, i.e., a manifold with a defined way of measuring angles and distances.

For our case study, we chose the shape space of \textit{differential coordinates}~\cite{vonTycowicz_ea2018}, whose mathematical details are summarized in App.~\ref{app:shape_space}. This space can be used when the objects under study are digitally given as triangular meshes in correspondence, the latter being a semantically motivated one-to-one relation of the (necessarily same number of) vertices. 

\subsection{Statistics on manifolds} \label{sec:statistics}

\textcolor{black}{Statistical analysis is the de-facto standard for sets of data. To apply it in non-Euclidean manifolds a generalization of methods from multivariate statistics is required.} The first steps in this direction were taken a long time ago, e.g., by Fr\'{e}chet, but recently there have been many advances. With increasing computational power, it is now possible to incorporate the nonlinear structure inherent in many types of data into our models of reality and still perform the computations in a reasonable amount of time. As a result, there has been and continues to be an increasing amount of research that has led to a new field often referred to as \textit{geometric statistics}.
\textcolor{black}{As we will develop a new statistical tool for the analysis of the shadow surface shapes in Sec.~\ref{sec:normalization}, we introduce the notions from geometric statistics that are necessary to define it in the following.\footnote{\textcolor{black}{Given data in a manifold we will always assume that all data points can be pairwise connected by a unique shortest path. This is the case when the data is sufficiently localized in the shape space.
For example, within every hemisphere of the 2 dimensional sphere any two points can be joined by a unique segment of a great arc.
Thus, without mentioning it again, we always assume that the data is localized in such a neighborhood.}} Mathematical background can be found in App.~\ref{app:math_detail}. Since our new method works in any Riemannian manifold, we formulate it in general terms. As part of our study of the shadow surfaces in Sec.~\ref{sec:results} we apply it in the shape space of differential coordinates.}

Let $M$ be a manifold. Since linear combinations of its elements need not lie in $M$ again, even the simplest statistical index, the mean, is \textit{not} well-defined. A Riemannian metric, i.e., a smoothly varying inner product $\langle \cdot, \cdot \rangle_p$ on the tangent spaces $T_pM$, $p \in M$, yields a solution: It allows to measure angles between tangent vectors from the same tangent space and induces a distance between any two points in $M$. The resulting distance function, in turn, gives rise to a notion of mean (see, e.g., \textcolor{black}{Ref.~}\cite[Sec.\ 2.2]{Pennec_ea2019_book}) and principal component analysis (PCA) (see, e.g., \textcolor{black}{Refs.}~\cite{Fletcher_ea2004, HuckemannEltzner2020}), as well as further essential statistical quantities and procedures.

Around every point $p \in M$ the manifold can locally be mapped one-to-one into the tangent space $T_pM$ under preservation of point-to-origin distances. The corresponding function is called Riemannian logarithm $\log_p$. Its inverse is the Riemannian exponential $\exp_p$. 
Because of this, $\log_p(q)$ can be interpreted as the difference between $q \in M$ and $p \in M$: It is the vector in $T_pM$ that points to $q$ and is as long as the distance between them. A major difference between curved Riemannian manifolds and Euclidean space appears when one wants to compare vectors (and thus the above differences) from tangent spaces at different points: In manifolds, we must explicitly transport one to the other along a curve; this operation is called \textit{parallel transport}. Whenever the space is curved, the result of the transport depends on the chosen path.

\begin{figure*}[t]
    \centering
    \includegraphics[width=1\textwidth]{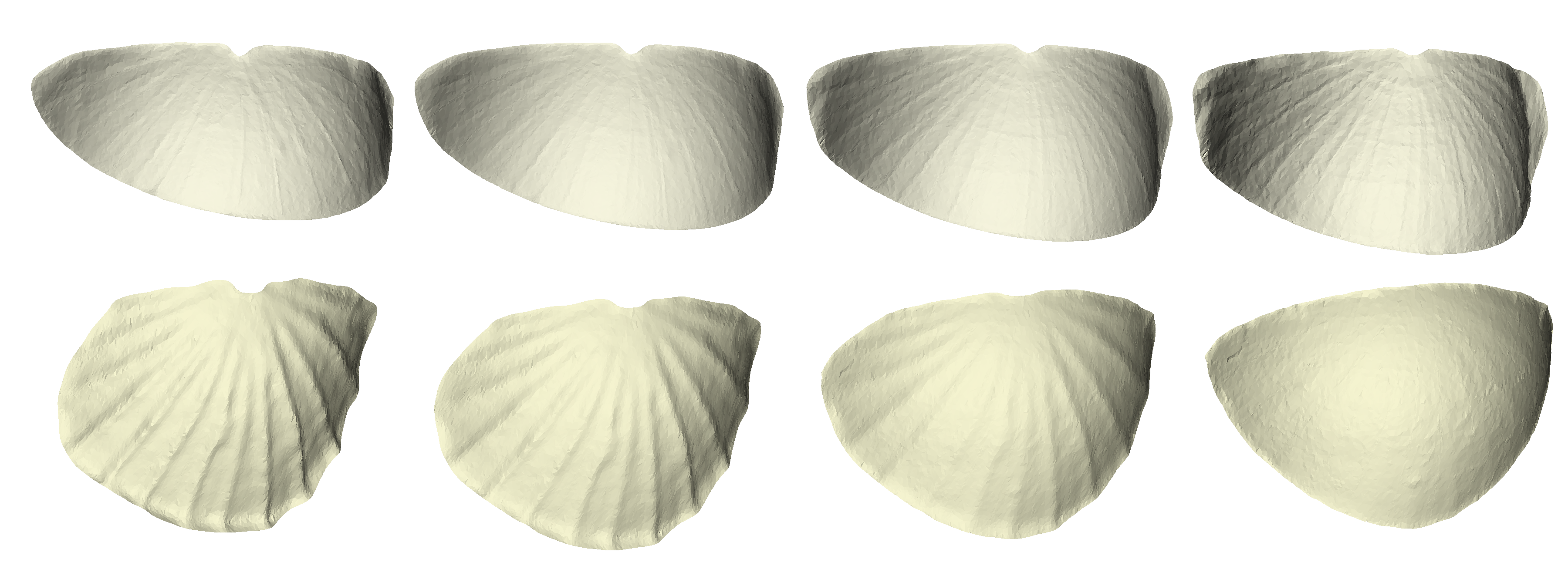}
    \caption{Results of geodesic regression for shadow surfaces of spherical sundials w.r.t.\ latitude. The results for Roman and Greek sundials are shown in the top and  bottom row, respectively. After calculating the trajectories in the space of differential coordinates, we sampled each curve at 4 points and computed the corresponding triangular meshes. The Roman geodesic models the trend for latitudes in $I_R = [40.7030, 43.3155]$ while the Greek one is defined on $I_G = [36.0917, 37.3900]$.}\label{fig:geodesics}
    \Description[Regression results]{Four samples of the curves obtained from regression w.r.t.\ latitude for both the Roman and Greek shadow surfaces}
\end{figure*}

Geodesics are very useful objects in Riemannian geometry. Being generalized straight lines in $M$, they are also (locally) shortest paths. In particular\footnote{In fact, geodesics play a major role in almost all concepts of geometric statistics, including the generalization of the mean and PCA, as they are the curves that realize the distance between points.}, they allow for the generalization of multivariate linear regression---a problem that can be understood as finding the best fitting straight line to a vector-space-valued data set. Analogously, in geodesic regression one tries to find the best fitting geodesic for manifold-valued data~\cite{Fletcher2013}. The latter method has mostly been used in biological and medical applications; however, in what follows we will show that it can be made fruitful for problems in archaeology as well. 
The mathematical details of geodesic regression can be found in App.~\ref{app:math_detail}; with Fig.~\ref{fig:geodesic_regression_synthetic} it also \textcolor{black}{yields} a visualization of the underlying idea. 
In our study we used this method to regress the shapes of the spherical shadow surfaces from each group (Roman and Greek origin) against the latitude of their site using the differential coordinates space; the result for spherical sundials are shown in Fig.~\ref{fig:geodesics}.

Finally, note that regression with generalized polynomials is also possible on manifolds~\cite{Hanik_ea2020}. We rely on geodesics in this article, but these higher order methods could be easily incorporated if a more complicated dependence on the parameter is expected.

\subsection{Regression-based normalization of data} \label{sec:normalization}
Let $M$ be a Riemannian manifold (e.g., a shape space) and $I_i \subset \mathbb{R}$, $i=1,\dots,k,$ closed intervals with non-empty intersection, i.e.,\ $\cap_{i=1}^k I_i \ne \emptyset$. 
The latter assumption guarantees a \textit{common} parameter value that serves as a reference point for the normalization.
We further assume that we are given $k$ groups of data points in $M$, where each element is labeled with a parameter: 
\begin{equation*}
\left( q^{(i)}_j, t^{(i)}_j \right) \in M \times I_i, \quad i=1,\dots,k, \quad j=1,\dots,n_i.
\end{equation*}
We assume that within each group the variability of the points $q^{(i)}_j$ with respect to the parameters $t^{(i)}_j$ can be well approximated by a geodesic.
Our goal is an inter-group comparison of the data without the confounding influence that is caused by different values of the parameters $t_i$. Therefore, we suggest to normalize the data in each group to the same value $t_0 \in \cap_{i=1}^k I_i$ using geodesic regression. Applying it separately to each group yields best-fitting geodesics $\gamma^{(i)}: I_i \to M$, $i = 1,\dots,k$. 
These geodesics can already provide valuable information about the data, as we will see when we compute them for sundials.

\begin{figure*}[ht]
\centering
    \centering
    \includegraphics[width=.85\textwidth, trim={0 0 0 1.3cm},clip]{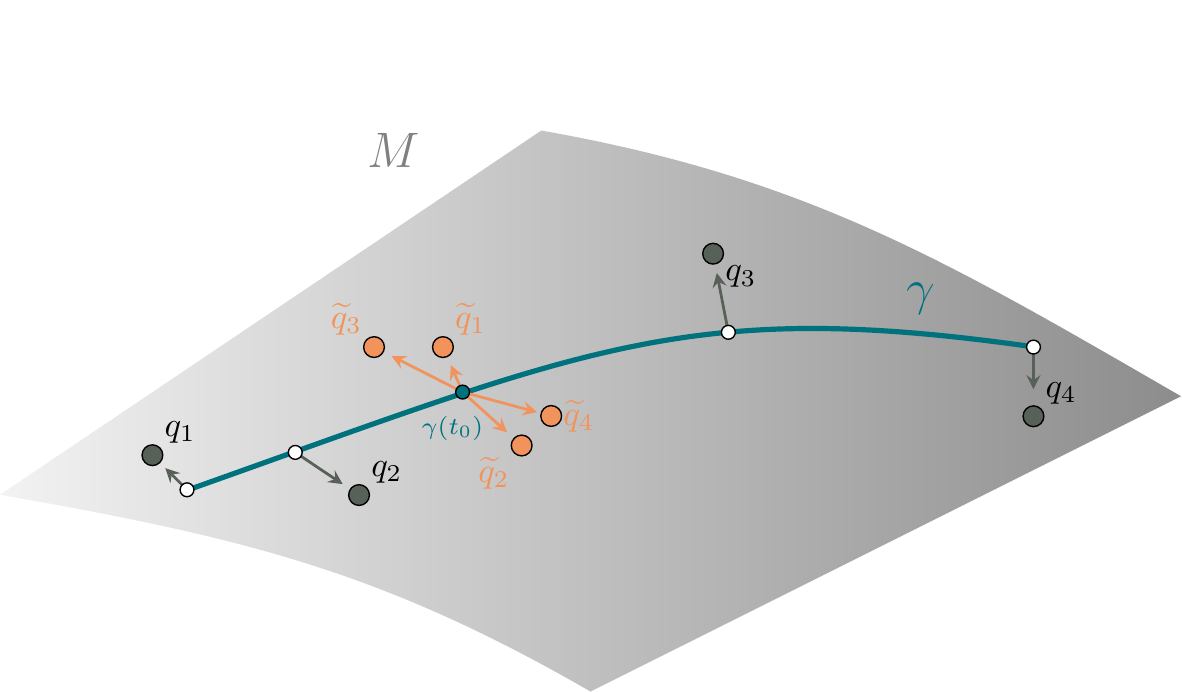}
    \caption{Normalization w.r.t.\ some parameter $t$ for a single data group $(q_j,t_j)$, $j=1,2,3,4$, in (a part of) a shape manifold $M$: Each point represents a complete shape. The curve $\gamma$ is the result of geodesic regression w.r.t.\ $t$. The points $\gamma(t_j)$ are depicted in white, while the tangent vectors $v_j$ are the gray arrows. Finally, the parallel translations $w_j$ of the $v_j$ in the tangent space $T_{\gamma(t_0)}M$ at $\gamma(t_0)$ are shown in orange, yielding the normalized data  $\widetilde{q}_j$.}\label{fig:normalization}
    \Description[Normalization schematic]{Visualization of data normalization with regression in an arbitrary Riemannian manifold}
\hfill
    \centering
    \includegraphics[width=0.85\textwidth, trim={0 0 0 1.3cm},clip]{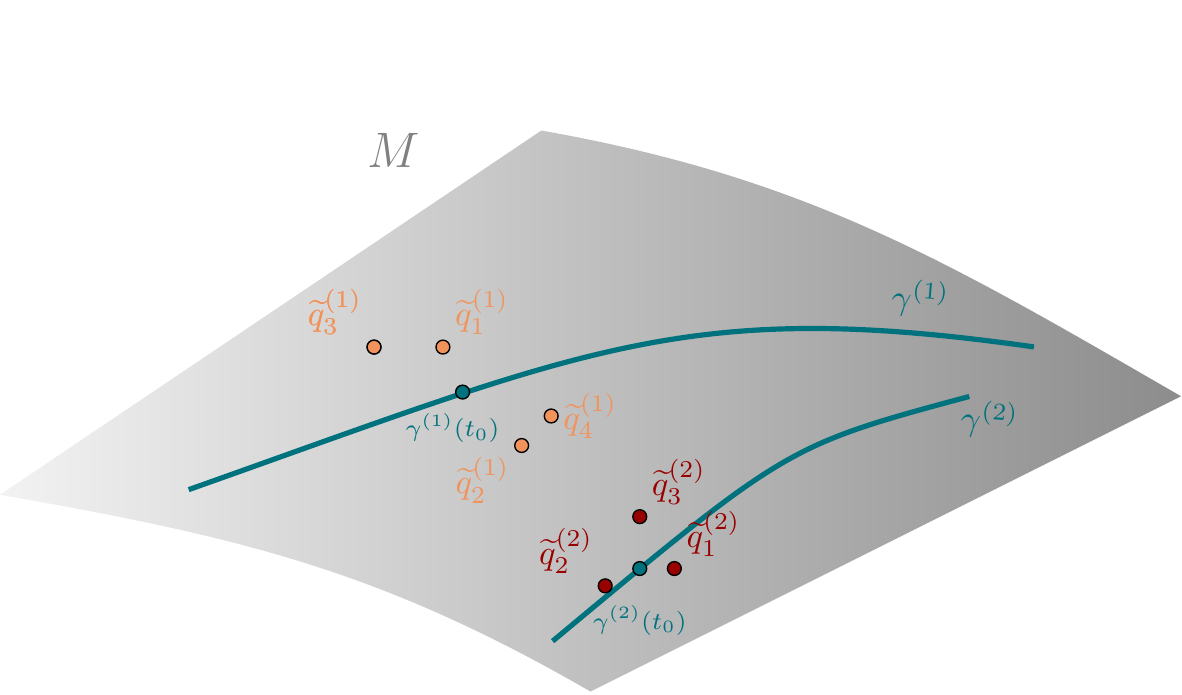}
    \caption{Normalization results for two groups: Both geodesics $\gamma_1$ and $\gamma_2$ are defined on the same interval and the data is normalized at the same value $t = t_0$. The groups (orange and red) can now be compared without a bias caused by the influence of \textcolor{black}{variable} $t$.}\label{fig:comparison}
    \Description[Group-wise comparison after normalization]{Comparison of data from 2 groups after normalization with regression in an arbitrary Riemannian manifold}
\end{figure*}

The difference of \textcolor{black}{each} point $q^{(i)}_j$ to its corresponding point $p^{(i)}_j := \gamma^{(i)} \left(t_j^{(i)} \right)$ on $\gamma^{(i)}$ can be encoded as the tangent vector 
$$v^{(i)}_j := \log_{p^{(i)}_j} \left( q^{(i)}_j \right) \in T_{p^{(i)}_j}M.$$ 
Using parallel transport, we can translate these vectors along $\gamma^{(i)}$ to $\gamma^{(i)}(t_0)$, resulting in vectors
$$w^{(i)}_j \in T_{\gamma^{(i)}(t_0)}M.$$
They represent the differences of the data points to the geodesics \textit{normalized at} $t_0$. Mapping them back to the manifold gives the normalized data points:
$$\widetilde{q}^{(i)}_j := \exp_{\gamma^{(i)}(t_0)} \left( w^{(i)}_j \right).$$

Now, we can perform \emph{inter-group comparison} at the normalized value $t_0$. For example, we can compute the Fr\'{e}chet mean and perform group tests for equality of them; for the latter see \textcolor{black}{Refs.}~\cite[Sec.\ 3.3]{MuralidharanFletcher2012} and \cite{HanikHegevonTycowicz2020}. The normalization process and its result for \textcolor{black}{the} data \textcolor{black}{of} two groups are visualized in Figs.~\ref{fig:normalization} and \ref{fig:comparison}. We also provide an example:

\begin{exmp} \label{exmp:1}
    Let $M$ be a shape space with manifold structure like the one introduced in Sec.~\ref{sec:shape_space}. Suppose that we are given two sets of knives, say from Greece and Spain, and let $p^{1}_1,\dots,p^{1}_m$ and $p^{2}_1,\dots,p^{2}_n$ be their encoded shapes in $M$, respectively. Further, suppose that we know their creation dates $t^{1}_1,\dots,t^{1}_n$ and $t^{2}_1,\dots,t^{2}_n$, which lie between the years 0 and 300 AC. 
    
    Now, assuming a continuous \textcolor{black}{evolution} of \textcolor{black}{knife shapes over} time, we can apply our method to compare them after removing the chronological trend\textcolor{black}{. F}or this, we set $I_1 = I_2 := [0,300]$\textcolor{black}{; t}hen, the shapes $\widetilde{q}^{(i)}_j$ approximate how the knives of both groups looked at any (fixed) time $t_0 \in [0, 300]$.
\end{exmp}

Note that geodesics can be extended in almost all practical scenarios (see \textcolor{black}{Ref.~}\cite[Ch. 7]{doCarmo1992} for details). (In ordinary Euclidean space, this corresponds to continuing a straight line.) This allows---within limits that depend on the data---to extrapolate the trends $\gamma^{i}$ so that they are defined on larger intervals. This can be very helpful when $\cap_{i=1}^k I_i = \emptyset$, but moderate extrapolation assures overlapping intervals. We used this property in our study to extend two intervals so that their intersection is non-empty.

\subsection{Analysis pipeline} \label{sec:pipeline}

\begin{figure*}[ht]
    \centering
        \includegraphics[width=\textwidth]{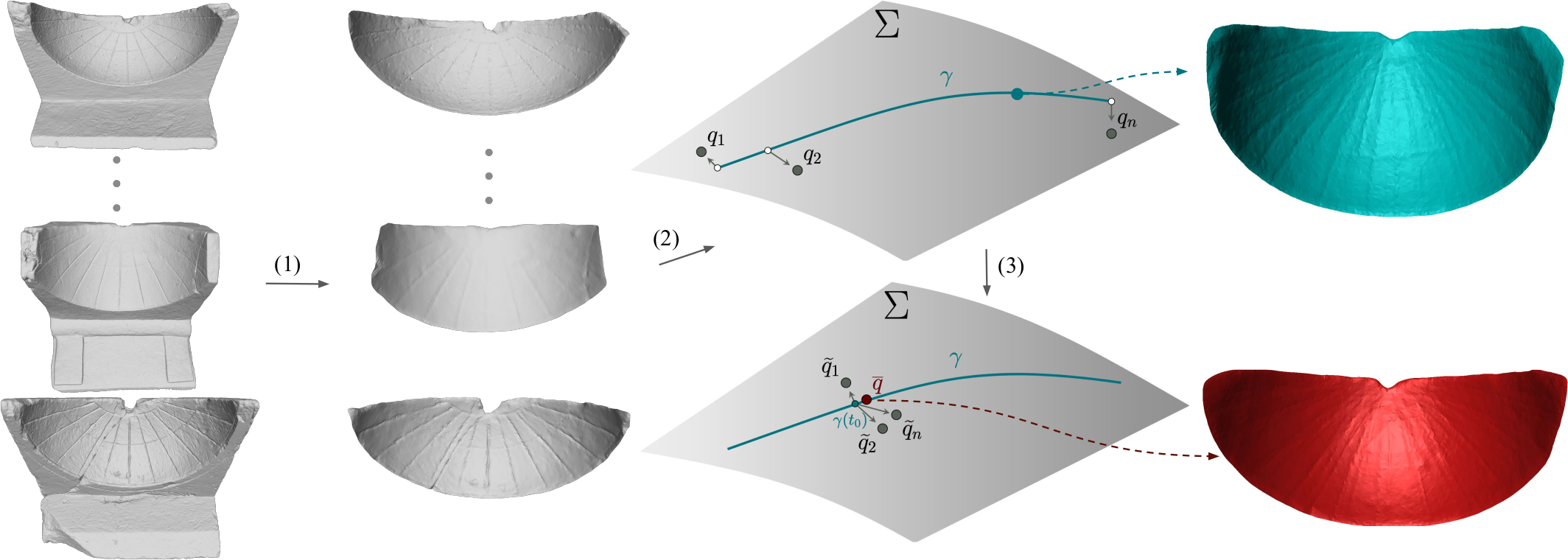}
    \caption{Visualization of the analysis pipeline.}\label{fig:pipeline}
    \Description[Pipeline]{Proposed pipeline for statistical analysis of sundial shape data}
\end{figure*}

We now discuss the analysis pipeline that we applied separately to the Roman and Greek groups; it is visualized in Fig.~\ref{fig:pipeline}. Let $n$ denote the number of sundials in the group. After digitizing the sundials in the form of triangle meshes, shadow surfaces were extracted and a group-wise correspondence was established (1). The shapes $q_1,\dots,q_n$ were then encoded in the shape space $\Sigma$ of differential coordinates (2). The statistical analysis of the shapes of the shadow surfaces was then performed in the manifold $\Sigma$; we applied geodesic regression \cite{Fletcher2013} w.r.t.\ latitude, which yielded the fitted geodesic $\gamma$ (as depicted). Being a function that depends on latitude, we could evaluate $\gamma$ at various latitudes. Any resulting shape represents a typical shadow surface at the given latitude. It can be transformed into a triangular mesh for visualization (the green shadow surface in Fig.~\ref{fig:pipeline}). 

The Roman geodesic already revealed interesting information: We used it to infer the latitude of the installation site of an unknown sundial (see next section). 

In step (3), the novel normalization method was applied to normalize all shapes to a common latitude $t_0$, resulting in points $\widetilde{q}_1,\dots,\widetilde{q}_n \in \Sigma$. The shapes of the shadow surfaces could now be further analyzed without the additional variability introduced by the influence of latitude. From these normalized shapes, we computed the mean shape $\overline{q}$ at $t_0$. A visualization of this mean (the red shadow surface in Fig.~\ref{fig:pipeline}) can again be obtained when we transform $\overline{q}$ back into a triangular mesh.

\section{Results}\label{sec:results}

\begin{figure*}[ht]
    \centering
        \includegraphics[width=.9\textwidth]{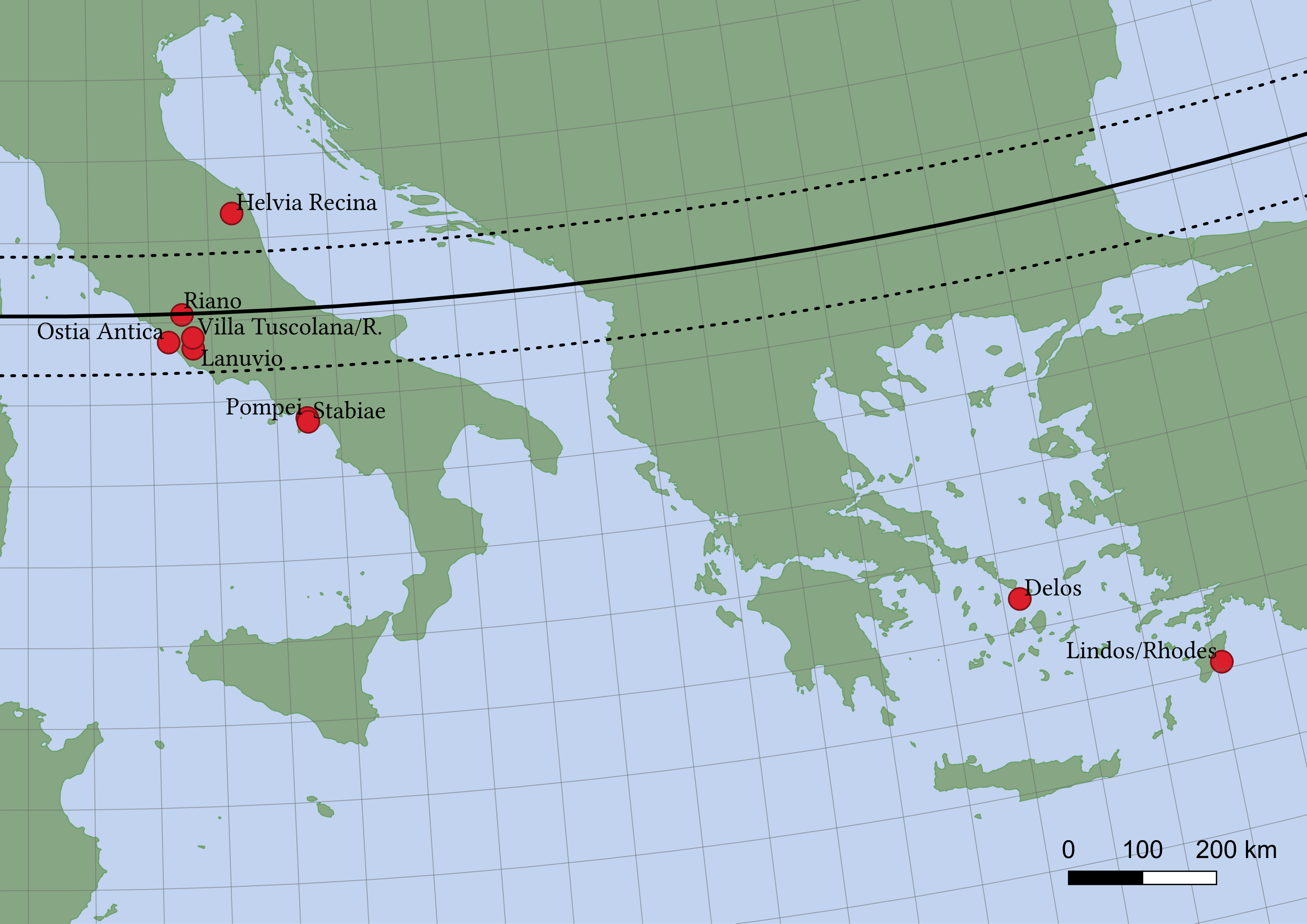}
    \caption{Map of the relevant area: The sites of the used sundials are given by the red circles. The solid line indicates the latitude at which the sundial with Dialface ID 39 was probably installed. The dashed lines bound the region of uncertainty of this prediction as determined by the computed MAE from all sundials with known location of installation.}\label{fig:map}
    \Description[Map with locations]{Map with locations of installation of the studied sundials including predicted latitude of ID39 and area of uncertainty}
\end{figure*}
In this section, we discuss the results of our case study. First, we describe how we used geodesic regression to analyze the latitude dependence of ancient Roman sundials from the Italian peninsula and how we used the results to narrow down the possible location of installation of a sundial with uncertain site. Then, we show an illustrative application of the normalization method from Sec.~\ref{sec:normalization} by comparing shadow surfaces from Greece with those from the Italian peninsula.

We used 10 Roman sundials from the Italian peninsula; their IDs as well as their sites---including longitude and latitude---can be found in App.~\ref{app:ID}; the latter are also depicted in Fig.~\ref{fig:map}. We pre-processed them as described in Sec.~\ref{sec:data}. 

\begin{figure*}[ht]
\centering
    \includegraphics[width=.95\textwidth]{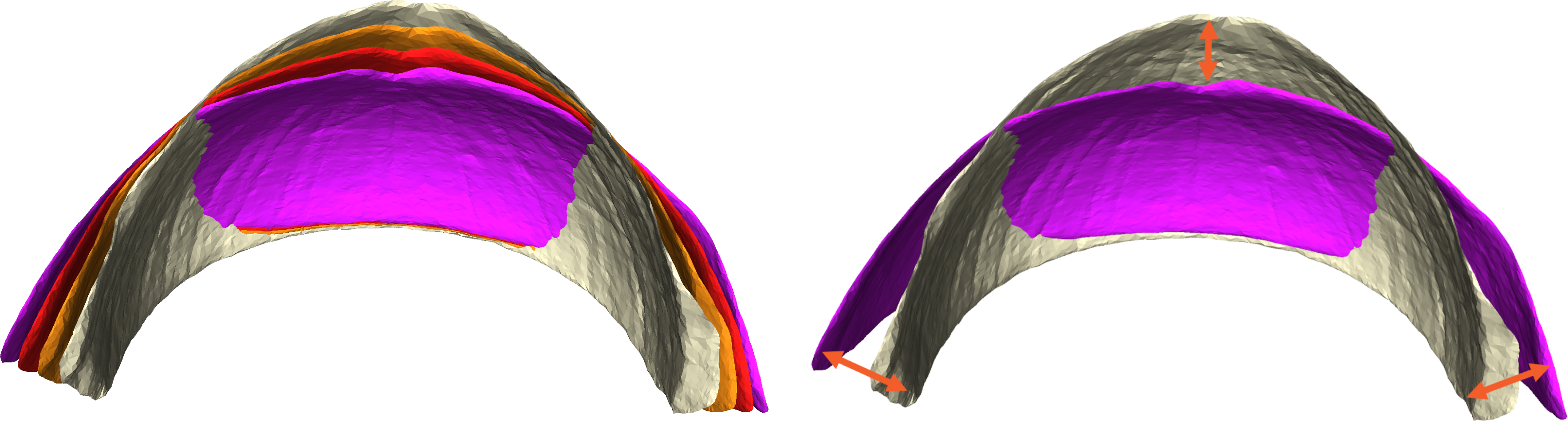}
\caption{Geodesic regression w.r.t.\ latitude over the interval $I_R = [40.7030, 43.3155]$ for shadow surfaces of Roman sundials. We evaluated the optimal geodesic at 4 equidistant points and show the top view of the resulting shadow surfaces, with colors beige, orange, red, and violet ranging (in this order) from the most northern to the most southern site. On the left, the shadow surfaces are overlaid to show the bending. The latter is also indicated by arrows on the right hand side, where only the shadow surfaces belonging to the most northern and southern latitude are shown.}\label{fig:geodesic_regressions}
\Description[Bending in Roman shadow surfaces]{Overlaid samples of the results from regression for Roman shadow surface at different latitudes that visualize the bending w.r.t.\ latitude}
\end{figure*}

    Given the shapes in terms of differential coordinates (i.e., points in shape space), we performed geodesic regression with the latitude of the site as explanatory variable;
    the corresponding interval was determined by the latitudes of the most southern and northern site, i.e., $I_R = [40.7030, 43.3155]$. It seemed reasonable to assume that a change in latitude causes a geodetic drift in the data: in the range of latitudes we studied, shapes can be expected to change at a constant rate in a fixed direction in shape space, to compensate for the latitude-dependent shift in sunlight angle.\footnote{To corroborate this, we also tested regression with higher-order curves in shape space, i.e., quadratic and cubic generalized B\'{e}zier curves, as explained in \textcolor{black}{Refs.}~\cite{Hanik_ea2020}. This overfitted the data, i.e., nearly interpolated the given shadow surfaces at their corresponding latitudes, \emph{while showing unreasonable behavior in between.}} Equidistant samples of the resulting geodesics are depicted in the top row of Fig.~\ref{fig:geodesics} and Fig.~\ref{fig:geodesic_regressions}. They show a ``bending'' of the shadow surface that increases with latitude; \textcolor{black}{this} is highlighted in Fig.~\ref{fig:geodesic_regressions}.

\begin{figure*}[ht!]
    \begin{center}
        \includegraphics[width=0.7\textwidth]{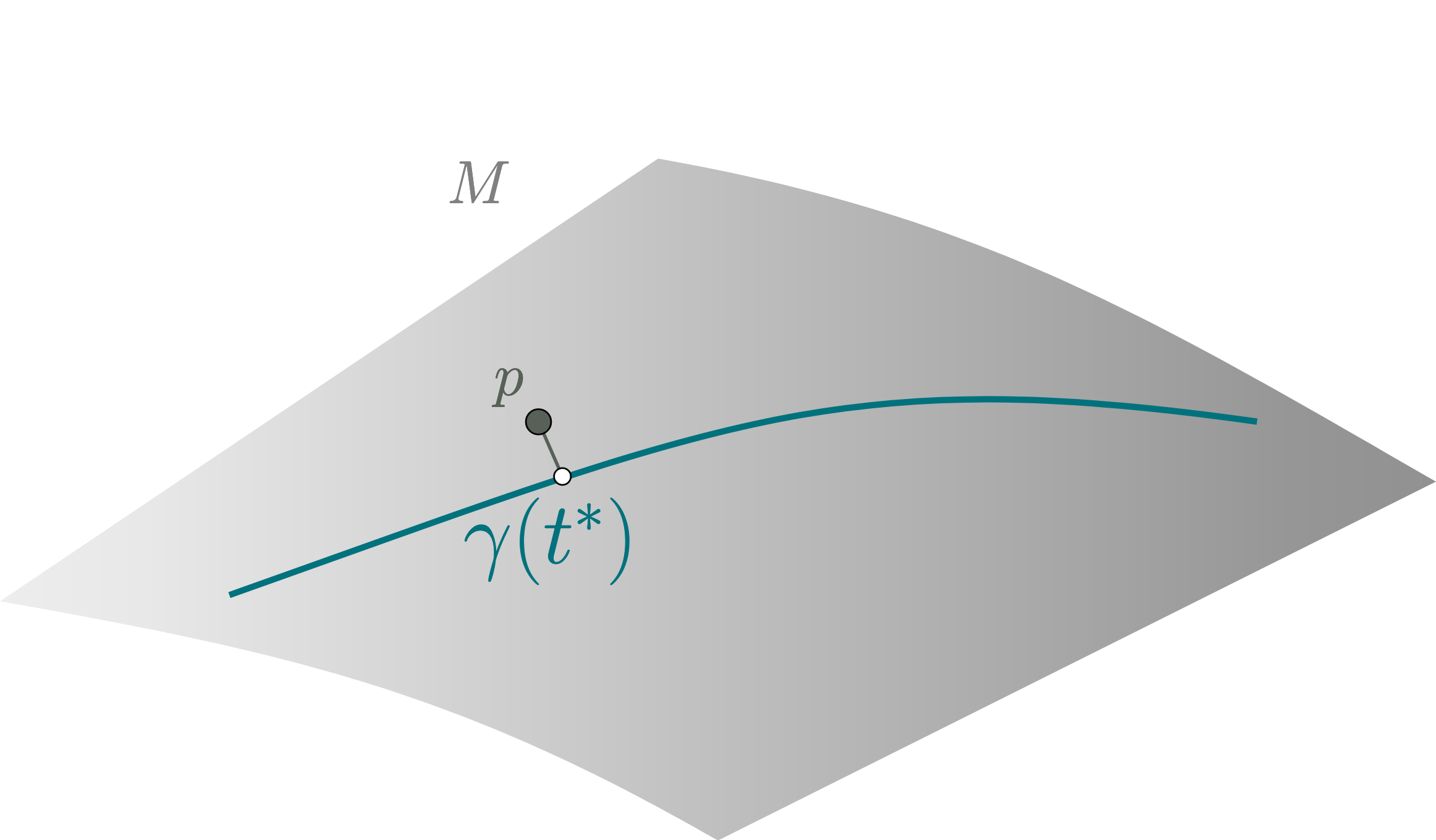}
    \end{center}
    \caption{\textcolor{black}{Projection of an arbitrary point $p$ in a manifold $M$ onto a geodesic $\gamma$. The number $t^*$ is the corresponding parameter of $\gamma$, i.e., the minimizer of $\mathcal{E}(t):= d(\gamma(t), p)$, where $d$ is the Riemannian distance in $M$.}}
    \Description[Projection]{Visualization of projection onto a geodesic in an arbitrary Riemannian manifold}
    \label{fig:projection}
\end{figure*}

The ``bending'' is an interesting effect: 
it suggests that the \emph{form} of the whole shadow surface (and not only that of the hour lines) were adapted to the location of installation\footnote{\textcolor{black}{It would be interesting to know how this was technically realized in production. However, we are not aware of any ancient source that provides information about this.}}.
This fact can be used to place sundials of unknown or uncertain installation site: 
after projecting orthogonally~\cite[Sec.\ 3.3]{AmbellanZachowvonTycowicz2021} the shape of the sundials' shadow surface onto the regressed geodesic (which here mostly amounts to finding the most similarly bend shadow surface on the curve), we can ``read off'' the latitude at which the sundial was probably installed \textcolor{black}{(see Fig.~\ref{fig:projection}).}

\begin{figure*}[ht!]
    \begin{center}
        \includegraphics[width=0.7\textwidth]{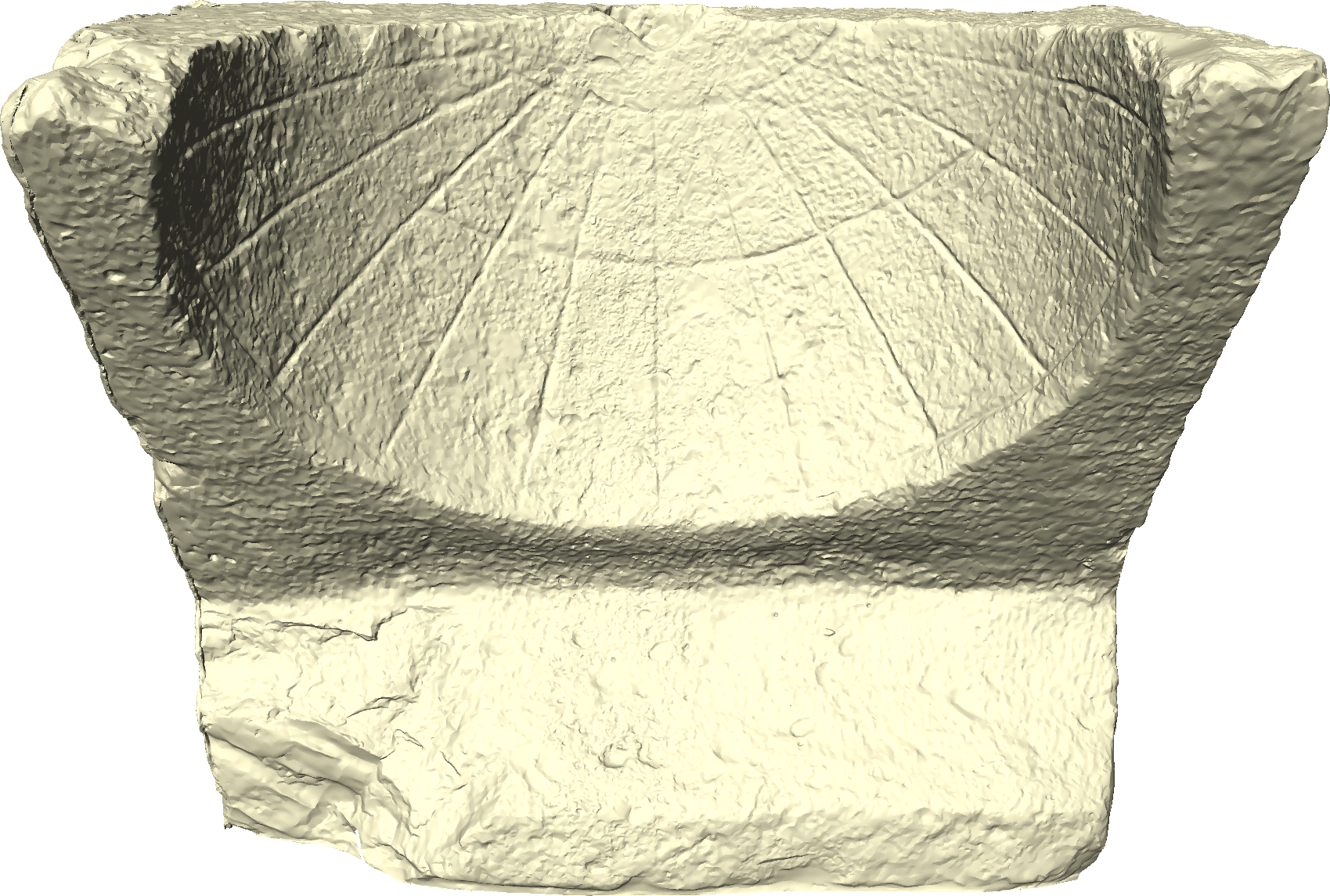}
    \end{center}
    \caption{3D model of the sundial with dialface ID 39.}
    \Description[Sundial ID39]{Image of sundial ID39}
    \label{fig:ID39}
\end{figure*}

To demonstrate this, we selected a sundial from Italy (Dialface ID 39 in the Topoi database; see Table~\ref{table2}), which is now in a museum in Vatican City, but whose installation site is \textcolor{black}{uncertain}. \textcolor{black}{(Fig.~\ref{fig:ID39} shows a visualization of this sundial.)}
Projecting the shape of its shadow surface onto the Roman geodesic gave an approximate latitude of $t^* \approx 42.1029$\textdegree, which is slightly north of the latitude of Rome. Thus, our model suggests that the sundial was once located at a site near the \textit{solid} line shown in Fig.~\ref{fig:map}.
In order to investigate the accuracy of this prediction we performed a leave-one-out cross validation test for the Roman sundials with \textit{known} location: Each sundial was taken out, regression was computed using all other samples, and the latitude of the left-out sample was predicted. The \textit{mean absolute error} (MAE) we thereby found was approximately \textcolor{black}{$0.75$\textdegree\ latitude (with standard deviation of $0.46$\textdegree)}, which is about 80 kilometers in north-south direction. 

\begin{figure*}[bt!]
    \begin{center}
        \includegraphics[width=0.9\textwidth]{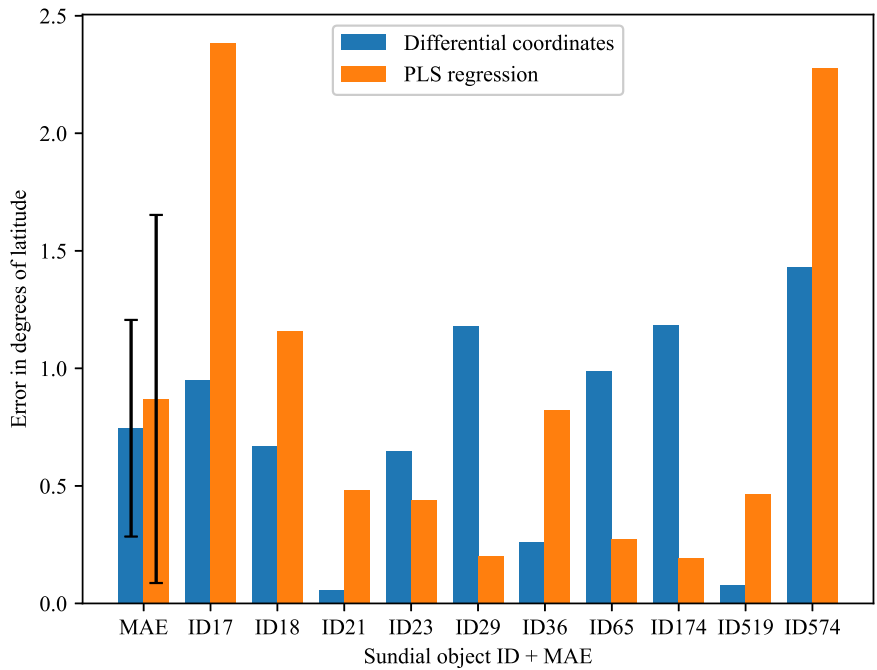}
    \end{center}
    \caption{\textcolor{black}{Sundial-wise and mean absolute errors obtained during latitude prediction. ``Differential coordinates'' refers to the proposed method; ``PLS regression'' stands for the results of the method that is based on partial least squares regression. MEAs are depicted with standard deviations.}}
    \Description[Bar plot of prediction errors]{A bar plot of the individual errors in latitude prediction in degrees of latitude}
    \label{fig:prediction_errors}
\end{figure*}

We use the computed MAE as a measure of uncertainty for the prediction of the unknown sundial and depict the latitudes \textcolor{black}{$t^* \pm 0.75$\textdegree\ } as \textit{dashed} lines in Fig.~\ref{fig:map}; it is likely that the location of installation was inside the resulting band.
\textcolor{black}{To compare our method against a traditional approach, we additionally applied \textit{partial least squares regression}~\cite{Wegelin2000} on the Procrustes aligned coordinates of the shadow surface meshes in order to predict the corresponding latitude. For this, we used the ``PLSRegression'' module from scikitlearn 1.1.1. Setting the number of components to $1$ yielded the best results. The MAE we thereby obtained was approximately $0.87$\textdegree\ latitude (with standard deviation of $0.78$\textdegree). This was about $0.12$\textdegree\ (about $12.32$ km) larger than the MAE of our proposed method. Furthermore, the standard deviation was substantially higher when using PLS regression. The sundial-wise errors and MAEs of both methods are shown in Fig.~\ref{fig:prediction_errors}.}

\bigskip

We used the normalization method to compare the mean shape of shadow surfaces from the Italian peninsula with that of 3 samples from Greece; information on the latter can be found in Table~\ref{table2} while their sites are also depicted in Fig.~\ref{fig:map}. Controlling for latitude before calculating the means correctly accounts for the fact that all Greek sundials were located further south than those from the Italian peninsula (an essential step at least for the Roman sundials).
Because the experiment could only be performed with 3 Greek samples, it should be seen as an illustrative example. Nevertheless, we discuss possible research questions that our experiment could answer 
\textcolor{black}{in the discussion section}.

\begin{figure*}[ht]
    \centering
    \includegraphics[width=.9\textwidth]{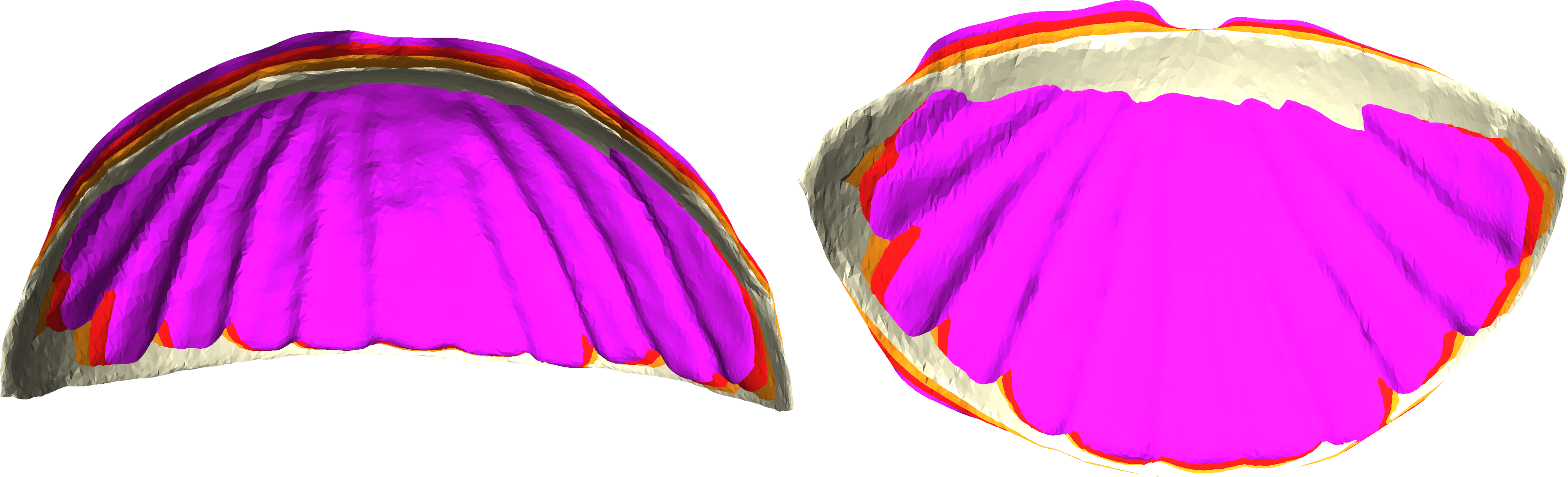}
    \caption{Geodesic regression w.r.t.\ latitude for shadow surfaces of sundials from Greece $I_G = [36.0917, 37.3900]$; the top view is on the left while the front view is on the right. We evaluated the resulting optimal geodesic at 4 equidistant points, with colors beige, orange, red, and violet ranging (in this order) from the most northern to the most southern site.} \label{fig:greek_geodesic}
    \Description[Regression for Greek shadow surfaces]{Overlaid samples of the result from regression w.r.t.\ latitude for Greek shadow surfaces.}
\end{figure*}

\begin{figure*}[ht]
    \centering
    \includegraphics[width=.9\textwidth]{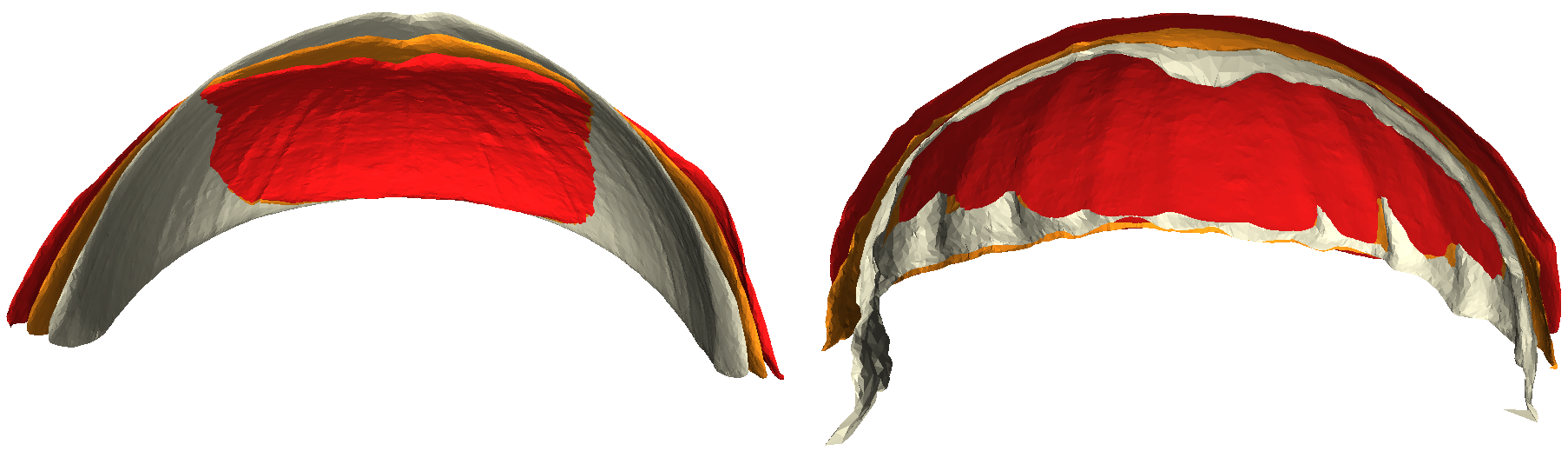}
    \caption{Roman (left) and Greek (right) means of shadow surfaces of spherical sundials after normalizing the Roman data at $40$\textdegree\ (red), $41$\textdegree\ (orange) and $42$\textdegree\ (beige) latitude and the Greek data at $37$\textdegree\ (red), $38$\textdegree\ (orange) and $39$\textdegree\ (beige) latitude, respectively.}\label{fig:means_40_41_42}
    \Description[Comparison of normalized means]{Overlaid normalized means at different latitudes within each group}
\end{figure*}

We performed geodesic regression with respect to latitude for the Greek sundials over the underlying interval $I_G = [36.0917, 37.3900]$; samples from the resulting geodesic are shown in Fig.~\ref{fig:greek_geodesic}. The shadow surface hardly bends with changing latitude contrasting our earlier result for Roman sundials. 
Since $I_R \cap I_G = \emptyset$, we extrapolated both the Roman and the Greek geodesic such that they were defined on the intervals $\widetilde{I}_R := [38.5,43.3155]$ and $\widetilde{I}_G := [36.0917, 39]$, respectively. This allowed us to normalize both data sets at different latitudes and, in each case, to compute the mean shape of the normalized shadow surfaces as described in Sec.~\ref{sec:normalization}; the results are shown in Fig.~\ref{fig:means_40_41_42}. Note that the means ``bend'' like the geodesics, which is the effect of the normalization.

Widening the window for comparison by further extrapolation was of no avail due to increased artifact formation in the normalized data.
The appearance of artifacts can be attributed to initial noise (\textcolor{black}{occurring mainly} at the corners, as explained in Sec.~\ref{sec:data}) that is amplified too much. First signs of such artifacts can be seen in Fig.~\ref{fig:means_40_41_42}\textcolor{black}{; they} begin to form at the corners of the Greek mean at $39$\textdegree\ latitude.  

\begin{figure*}[ht]
\centering
    \includegraphics[width=1\textwidth]{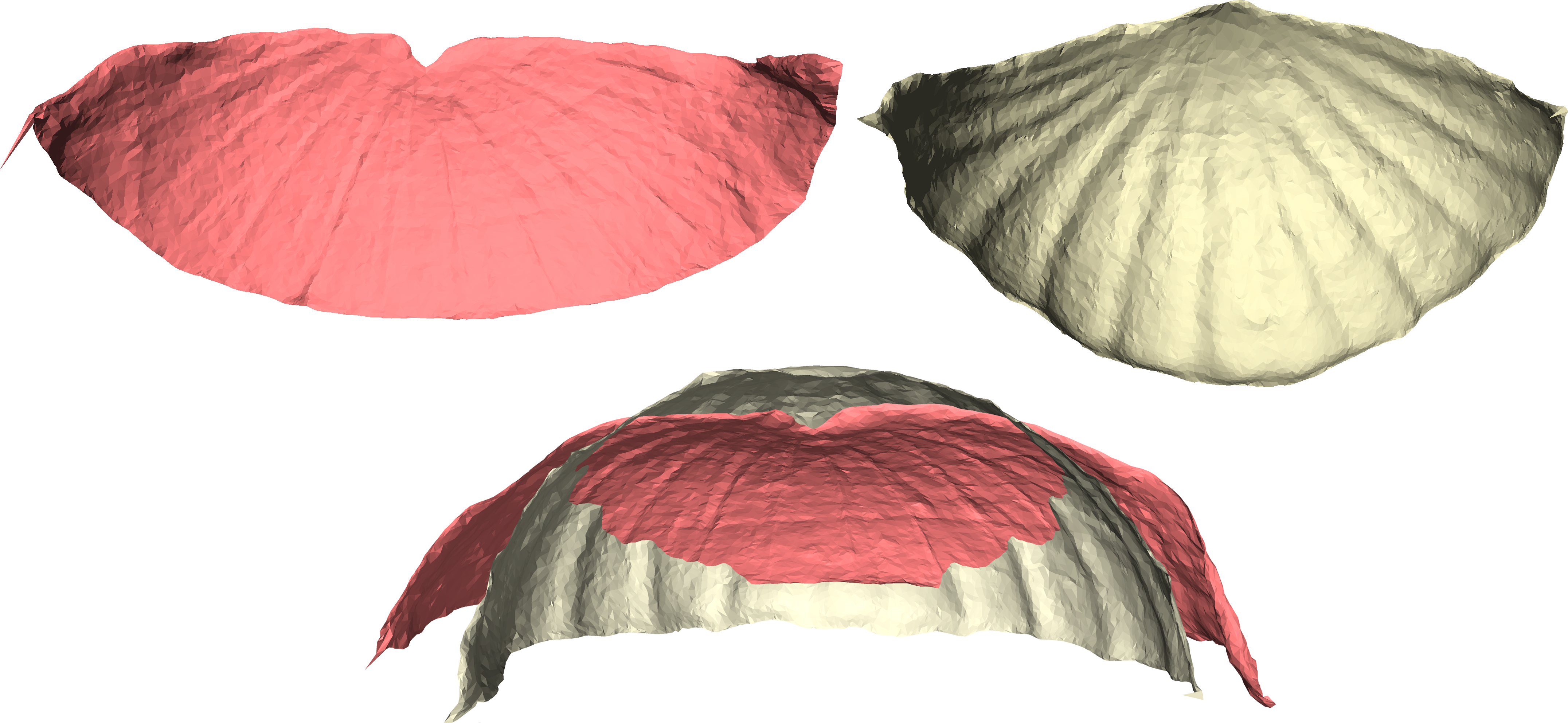}
\caption{Comparison of the Roman (red) and Greek (beige) mean shapes after normalizing \textcolor{black}{the respective samples with the proposed method} at $38.5 $\textdegree. They are shown both individually (top row) and overlaid (bottom row).}\label{fig:comparison_means}
\Description[Comparison of normalized group means]{Comparison (both individually and overlaid) of Roman and Greek mean shadow surface normalized at 38.5 degrees latitude}
\end{figure*}

\begin{figure*}[ht]
    \centering
    \includegraphics[width=.8\textwidth]{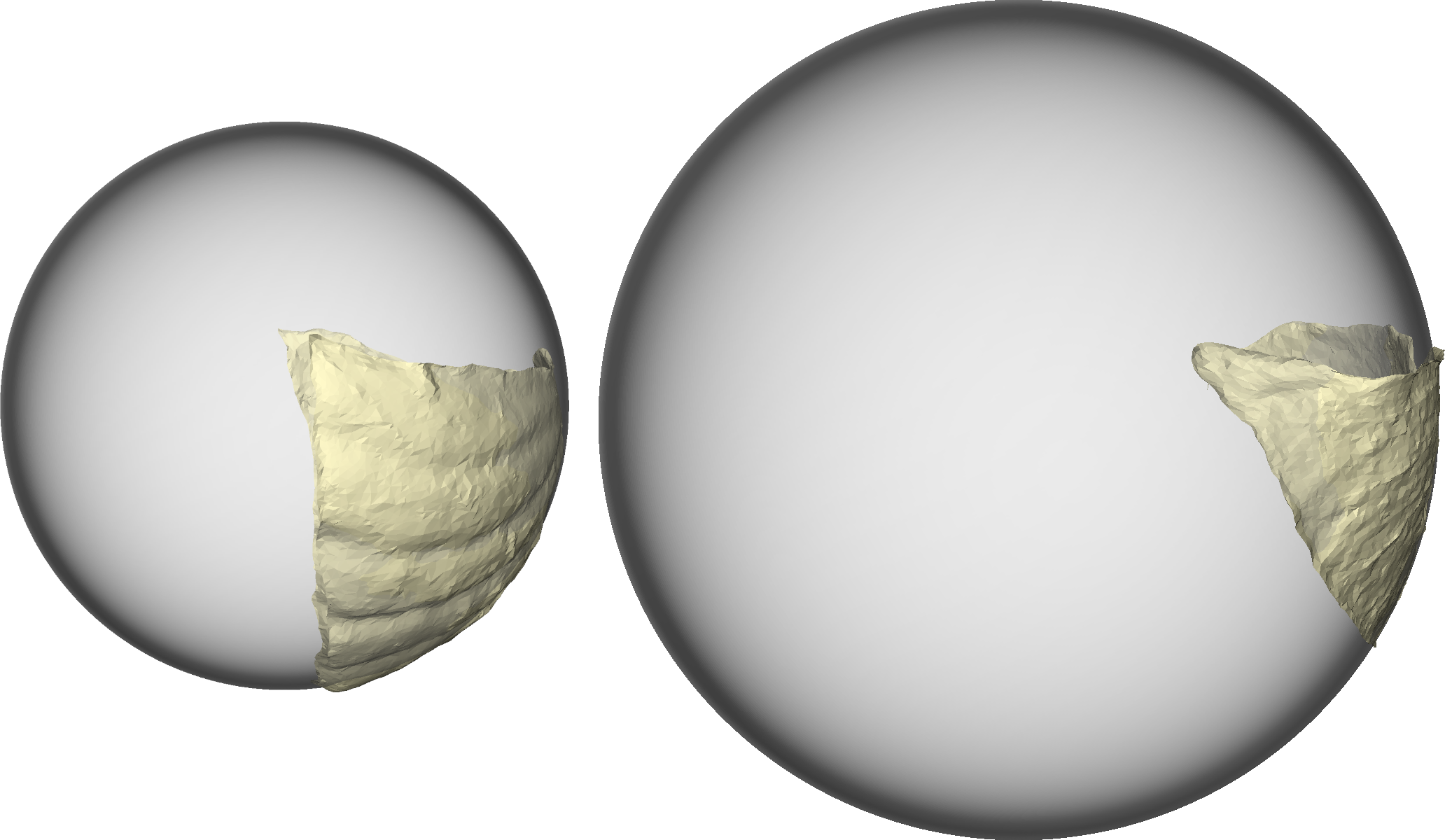}
    \caption{Spheres fitted to the Roman (left) and Greek (right) mean of the data that was normalized at $38.5$\textdegree\ latitude. \textcolor{black}{Center and radius of the spheres were chosen such that the average squared distances to the vertices of the triangle meshes were minimal.}}\label{fig:fitting_spheres}
    \Description[Comparison of inscribed spheres]{Comparison of inscribed spheres between the Roman and Greek mean of the normalized data}
\end{figure*}

In Fig.~\ref{fig:comparison_means}, we show both means after normalizing at $38.5$\textdegree\ latitude. When overlaid they look quite different---in particular their curvatures. (The artifacts at the corners are again due to extrapolation of noise.) In order to quantify this, we computed the spheres that fit the shadow surfaces best\footnote{\textcolor{black}{More precisely, we computed the center $c^* \in \mathbb{R}^3$ and radius $r^* > 0$ of the sphere with smallest sum of squared distances to the vertices of the shadow surface. Denoting the usual Euclidean norm by $\| \cdot \|$, it can be seen that the distance of a point $p \in \mathbb{R}^3$ to a sphere with center $c$ and radius $r$ is equal to $\tau(c,r; p) := | \|p-c\| - r|$. We thus computed $(c^*, r^*) := \min_{(c, r) \in \mathbb{R}^3 \times \mathbb{R}_{>0}} \sum_i\tau(c,r, v_i)^2$, where the $v_i$ are the vertices of the mesh of the shadow surface. The ``L-BFGS-B'' method was used for minimization. The problem is discussed in depth in Ref.~\cite{BermanCulpin1986}.}}; the results can be seen in Fig.~\ref{fig:fitting_spheres}. Both were different in size, the Greek sphere's radius measuring $66\%$ of the Roman. (Since size is normalized in our study only relative notions are of interest.) 
The fact that the ``bending'' is different influences the comparison of the means. 
As shown, at $38.5$\textdegree\ latitude they are differently curved; because of the stronger bending in the Roman group the forms would converge, though, if we used this trend to compare them further north.

In order to compare the non-linear shape space with a Euclidean alternative (which is difficult as no statistically significant quantitative results can be expected because of the small sample size) we investigated how the Mahalanobis distance between the mean shapes of the Roman and Greek groups adapts to the normalization. After mapping the Roman \textit{samples} and the Greek \textit{mean} to the tangent space at the Roman \textit{mean}, we computed the Mahalanobis distance\footnote{See \textcolor{black}{Ref.}~\cite[Sec.\ 7]{Pennec2006} for the Mahalanobis distance in manifolds. As the dimension of the differential coordinate space is (much) higher than the number of samples, the covariance matrix was not invertible. Therefore, as is often done, we substituted the inverse of the covariance matrix of the Roman samples by its pseudoinverse to compute the distance.} of the Greek mean to the distribution of the Roman samples. 
(We chose this direction because of the numbers of samples.) This was done for both the original data and the one normalized at 38.5\textdegree. Afterwards, the same computations were performed using the standard Euclidean shape space from~\cite{Cootes_ea1995}. For both shape spaces the distance increased when the samples were normalized. This makes sense because the in-group variability (here within the Roman group) reduces through the normalization (cf.\ Figs.~\ref{fig:normalization} and \ref{fig:comparison}). As the Mahalanobis distance weights the difference between the means (i.e., the logarithm) against the inverse of the observed covariance (matrix) in the Roman group, a decrease of the latter leads to a larger value. Interestingly, while the distance for the nonlinear space increased by a factor of $1.73$, it was multiplied by $5.1\times 10^5$ for the Euclidean model.\footnote{We also investigated how the geometric distance between the means behaves under the normalization. It increased by a factor of approximately $1.86$ and $1.64$ for the differential coordinate space and the Euclidean shape space, respectively. Therefore, the extreme increase in Mahalanobis distance for the latter cannot be explained by diverging means.} (Both numbers are rounded.) That is, the latter loses a high amount of variability while normalizing. We think that the differential coordinates conserve the finer details, whereas the Euclidean model tends to focus more on global features. But we believe that for the sundials smaller details are of high interest, in particular, when additional characteristics like the hour lines would be investigated as part of further analysis.

\begin{figure*}[!htb]
\centering
    \includegraphics[width=.8\textwidth]{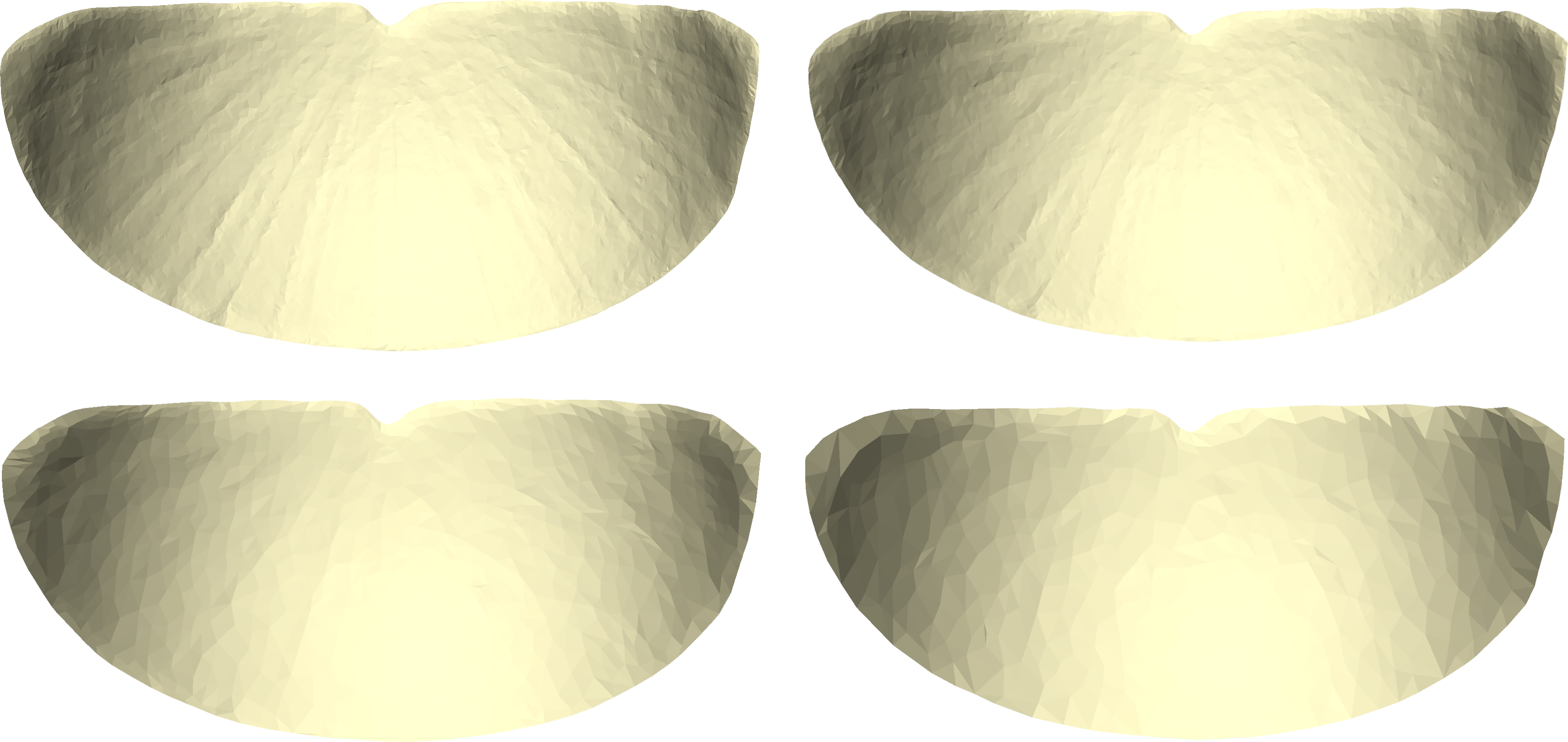}
\caption{Mean shapes of the normalized (at $41$\textdegree) sundials from the Italian peninsula for different mesh resolutions. Results are shown for resolutions of 20k (upper left), 10k (upper right), 2k (lower left) and 1k (lower right) faces.}\label{fig:simplification_italy}
\Description[Resolution comparison Roman]{Comparison of normalized means of the Roman shadow surfaces when computed with different resolutions shows little difference in results}
\end{figure*}

\begin{figure*}[!htb]
\centering
    \includegraphics[width=.8\textwidth]{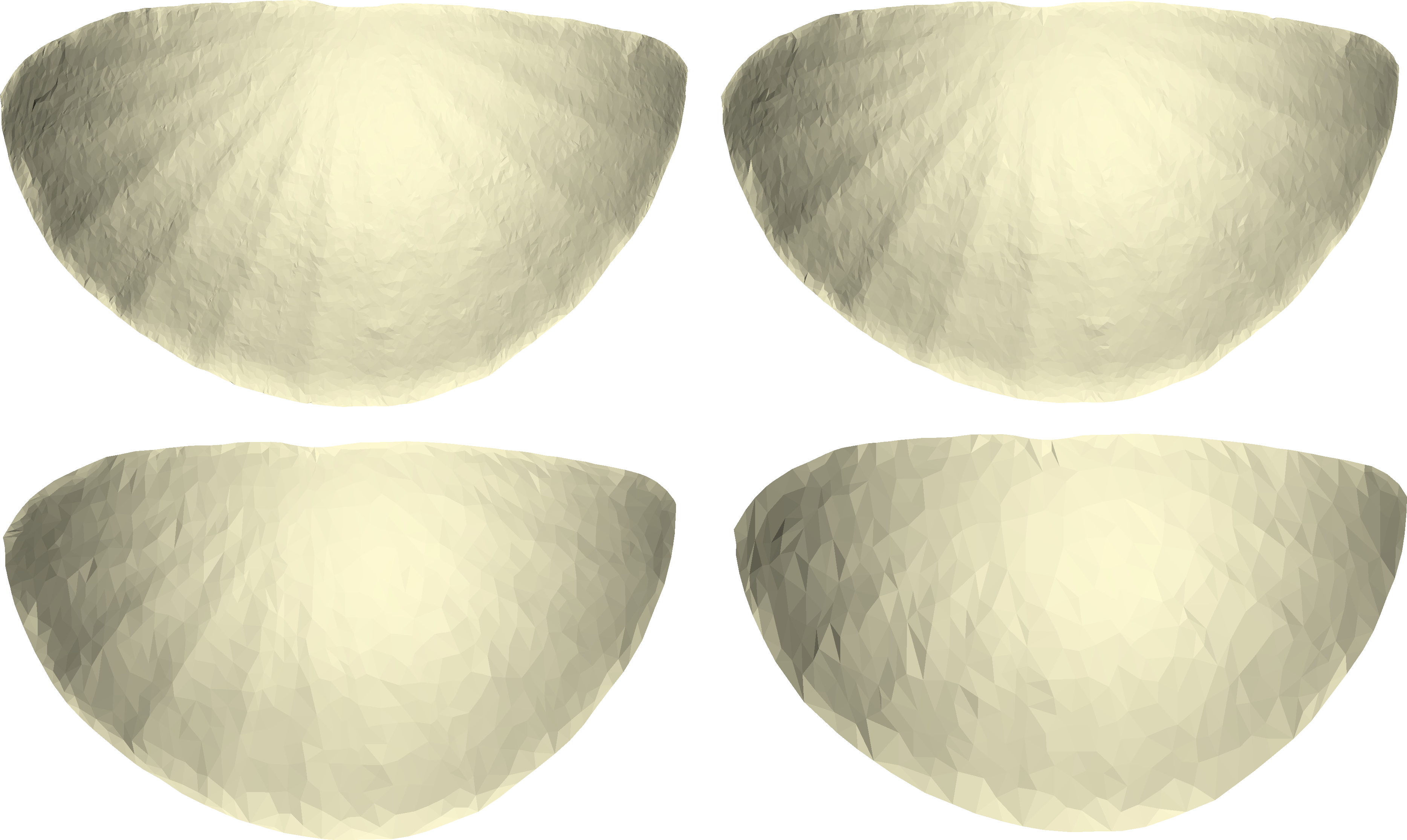}
\caption{Mean shapes of the normalized (at $37$\textdegree) sundials from Greece for different mesh resolutions. Results are shown for resolutions of 20k (upper left), 10k (upper right), 2k (lower left) and 1k (lower right) faces.}\label{fig:simplification_greece}
\Description[Resolution comparison Greek]{Comparison of normalized means of the Greek shadow surfaces when computed with different resolutions shows little difference in results}
\end{figure*}

\begin{figure*}[!htb]
    \centering
    \includegraphics[width=1\textwidth]{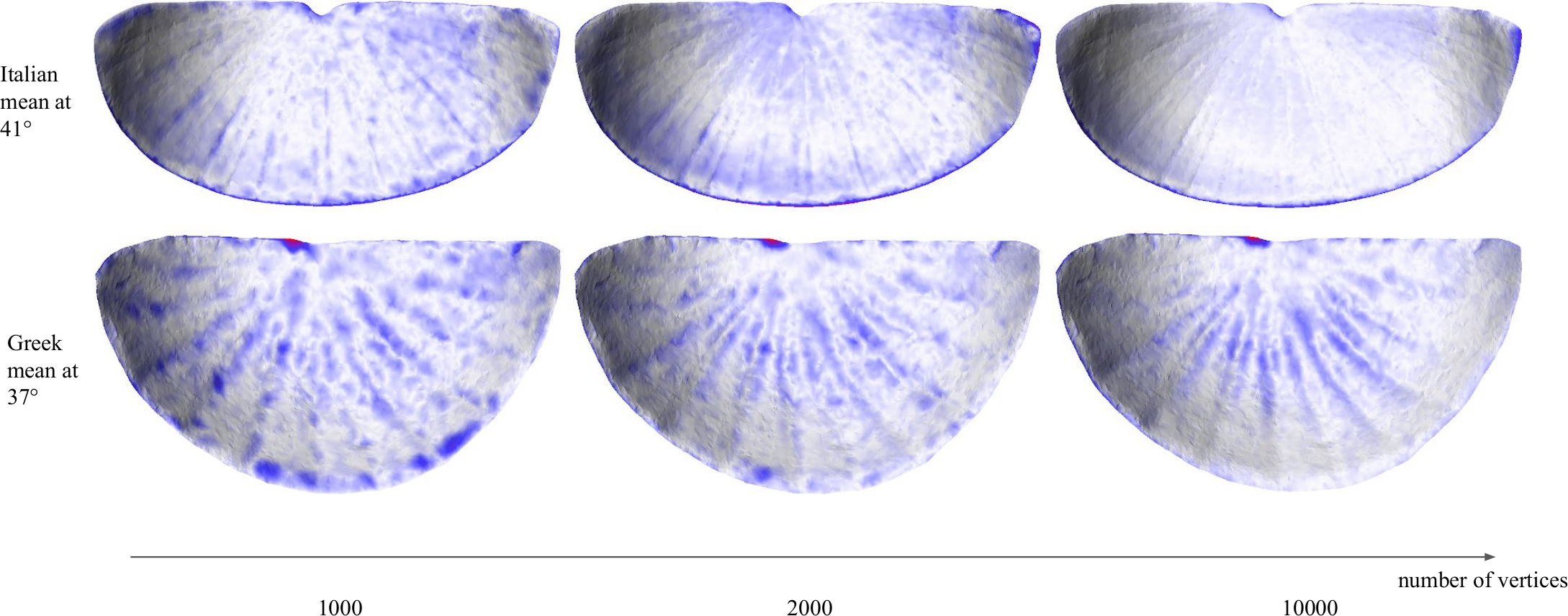}
    \caption{\textcolor{black}{Surface distance for different mesh resolutions when computing normalized means. The respective meshes of highest resolution are depicted with coloring (0.001 mm~\ShowColormap~5 mm) that encodes the distance to the corresponding mesh with lower resolution. Areas where the distance is smaller than 0.001 mm are white. To put this into context, the distance between the upper left and upper right corner were about 29 and 21 cm for the Italian and Greek mean, respectively.}}
    \label{fig:robustness}
\end{figure*}

We also examined to which extent the results depend on the resolution of the meshes by simplifying them. More precisely, we reduced the number of faces of each of the shadow surfaces (further) to 10k, 2k and 1k triangles with Meshlab's ``quadric edge collapse'' and computed normalized means for both the Roman and Greek group. As can be seen in Figs.~\ref{fig:simplification_italy}, \ref{fig:robustness}, \textcolor{black}{and~\ref{fig:simplification_greece}}, the means were effectively unchanged even with a poor resolution of only 1,000 faces. 


\color{black}

\section{Discussion}
The fact that the shape of the Roman shadow surfaces ``bends'' with latitude is a significant finding of this study. To the best of our knowledge it has never been observed before. It seems likely that the form of the whole shadow surface was adapted to the location of installation---a conclusions that is supported by the accuracy of the derivative method for latitude prediction. Our results thus add a piece to the puzzle how ancient Roman craftsman created working sundials. Clearly, an interesting question for future research is whether similar bending can also be observed for sundials from locations outside of Italy.

Our results also indicate that the new method to identify the latitude of the location of installation is accurate; furthermore, they suggest that utilizing the non-Euclidean shape space structure for prediction is superior to the commonly employed PLS regression. The only other method for the prediction of a sundial's working latitude we are aware of is the one from Pistellato, Traviglia, and Bergamasco~\cite{PistellatoTravigliaBergamasco2020}. Their method predicts the working latitude by utilizing the positions of the shadow surface's inscribed hour lines. Knowledge of celestial mechanics, which they assume the craftsmen possessed, allows them to compute the latitude from these positions. While their method has the advantage that it does not need additional sundials for prediction, it requires high mesh resolutions for accurate computations. In contrast, the above results show that the newly presented method works well even at coarse resolutions. Moreover, being purely data-driven, it requires no assumptions on the interplay between celestial mechanics and sundial shape. On the other hand, since both approaches make use of different, non-overlapping features of the shadow surface's make-up, they can  complement each other and even be applied together to utilize more of the available information.

A limiting factor of this study is the small number of samples that were available. Convincing statistical significance can only be achieved with more samples. This is a common situation in archaeology. While more ancient (digitized) sundials are available, they are too damaged to be used with current methods. New methods for statistical analysis of partial forms are needed to alleviate this problem; these could allow a greater number of existing ancient sundials to be used.


More data would also help to decide an interesting question \textcolor{black}{raised by the} proposed normalization method: whether craftsmen from Greece and the Italian peninsula, when building sundials, made different adaptions in order to account for changes in the angle of the sun. 
If backed up, the results of our experiment could \textcolor{black}{indicate} not only that the mean shape of a shadow surface depends on the latitude of the \textcolor{black}{installation site} in both regions, but also that the strength of dependence \emph{varies among the groups}.

It could be, for example, that the Greek craftsmen mainly adjusted the hour lines, while their Roman counterparts also adjusted the shape of the shadow area. It seems very interesting to investigate this question further; however, it would require more data, especially from Greece.

Last but not least, we clearly showed that the \textcolor{black}{proposed} methods \textcolor{black}{can be used even when} only relatively poor digital representations are at hand.



\color{black}

\section{Conclusion}\label{sec:conclusion}

Recent advances in image-based reconstruction (Structure from Motion and Multi-view Stereo) have made the acquisition of highly detailed 3D artifact models simple, fast and economical. The sundial models \textcolor{black}{explored in the present study} represent a rapidly growing volume of 3D research data\textcolor{black}{. These data should be explored with mathematical approaches that take full advantage of the rich information they contain and avoid information losses such as those caused by \textit{a priori} typological grouping.} 
We demonstrated \textcolor{black}{such a procedure} combining shape space methods and statistics on manifolds, and showed that there is great potential \textcolor{black}{to} extract more information from \textcolor{black}{groups of} artifact shapes \textcolor{black}{using such methods} without prior assumptions.
Conversely, methods such as those discussed here have the potential to generate new, less biased and less convoluted baselines for typological seriation, in which the ordering of data represents a better defined (e.g., geographical or temporal) trend. \textcolor{black}{An obvious next step is thus to apply the proposed methods to other artifact collections and to explore their full potential in archaeology.}

\textcolor{black}{In this study, the proposed methods revealed a latitudinal dependency of the shape of shadow surfaces in sundials of the Italian peninsula. This sheds new light on the construction principles of sundials in ancient times. We were also able to show how this dependence can be used to accurately determine the latitude of a sundial's working location.}

\textcolor{black}{Shape spaces are a current research topic, including how best to deal with challenges that are particularly common in archaeological investigations, such as incompleteness and weathering of artifacts.
The majority of the considered shape spaces exhibits a Riemannian structure and, hence, can be combined with the presented approach. Because of this flexibility, we expect the proposed methods will be of great importance in a wide range of application scenarios.}

\textcolor{black}{Finally}, it is also worth noting that we introduced a new method that allows to control for confounding effects \textcolor{black}{caused by} parameter-dependent variations. \textcolor{black}{This method has practical uses in archaeology, as we have demonstrated. However, as a} generic \textcolor{black}{method, it can also} be applied to \textcolor{black}{any other} manifold-valued data. \textcolor{black}{Thus, this work is} a contribution not only to archaeology, but also to geometric statistics \textcolor{black}{in general}.



\appendix

\section*{APPENDIX}

\section{List of sundials}
\label{app:ID}

The sundials used in the study are listed in Table \ref{table2} along with the latitude and longitude of their sites. They are part of the Topoi database. The triangle meshes of the whole sundials can be downloaded from \url{http://repository.edition-topoi.org/collection/BSDP}. The metadata can also be found there. The pre-processed meshes of the shadow surfaces are also available for download \cite{HanikvonTycowicz2022}.  

\section{Shape space of differential coordinates}
\label{app:shape_space}
\textcolor{black}{In this section we recall the fundamental facts on the shape space of differential coordinates.
The underlying idea is to model a shape as the derivative of the deformation that a common reference mesh has to undergo in order to coincide with the object. 
Being a derivative, this representation is invariant under translations and, when the objects are scaled to the same size and Procrustes-aligned~\cite{Goodall1991} beforehand, effects of scale, rotation, and relative translation are also minimized. Conversely, to given differential coordinates a corresponding triangle mesh can be computed. This is used in many illustrations in this article.}

Suppose that homogeneous objects are given in the form of triangle meshes $T_i \subset \mathbb{R}^3$ that are in correspondence, scaled to the same size and Procrustes-aligned~\cite{Goodall1991}. Then, we can encode their shape in the space of differential coordinates as introduced in \textcolor{black}{Ref.}~\cite{vonTycowicz_ea2018}. To this end, we view each mesh $T_i$ as the result of a deformation $\phi_i$ of a common reference $\overline{T}$; i.e., each $\phi_i: \overline{T} \to T_i$ is an orientation-preserving, simplicial isomorphism that yields a semantic correspondence. We compute $\overline{T}$ in an iterative process as the mesh representation of the data's Fr\'{e}chet mean as follows. To initialize, we choose one of the given geometries as reference, say $T_1$, and encode the shapes of every $T_i$ as explained below. Then, we compute their mean shape (see  App.~\ref{app:math_detail}) and use its mesh representation to update the reference. This procedure is repeated until convergence, or until the norm of the update does not decrease any further. The method has the advantage that it minimizes the bias that is introduced by a particular choice of reference.

From here on, everything works the same way for all meshes $T_i$. For this reason, we omit the index. 

Let $m$ be the number of faces of $T$ (which does not depend on $i$ because of the one-to-one correspondence) and denote the set of (real) 3-by-3 matrices with positive determinant by $\Glp(3)$. 
The Jacobian $D \phi$ of $\phi$ is a constant 3-by-3 matrix $G_j$ with $\textnormal{det}(G_j) >0$ on each face $F_j$ of $\overline{T}$, i.e., 
$$D \phi \big|_{F_j} = G_j \in \Glp(3) \quad \text{ for all } j=1,\dots,m.$$
Hence, using the polar decomposition, we can find rotation matrices $R_j \in \SO(3)$ and symmetric positive definite matrices $U_j \in \Symp(3)$ such that 
$$G_j = R_j U_j\quad \text{ for all } j=1,\dots,m.$$
Both have a physical interpretation since $R_j$ and $U_j$ can be viewed as the rotation and stretch, respectively, that $F_j$ undergoes when $\overline{T}$ is deformed by $\phi$.
We call 
$$S := ((R_1,U_1),\dots,(R_m,U_m))$$ 
the \textit{differential coordinates of} $T$. Thus, the \textit{space of differential coordinates for triangle meshes with $m$ faces} is 
$$\Sigma_m := (\SO(3) \times \Symp(3))^m.$$
This is a manifold that can be endowed with a Riemannian product metric using the canonical metric of $\SO(3)$ and the Log-Euclidean metric~\cite{Arsigny_ea2007} of $\Symp(3)$. Using it, allows for efficient statistics in the space; see App.~\ref{app:math_detail} for a very brief introduction to this.
Moreover, given differential coordinates $S \in \Sigma_m$ and a reference $\overline{T}$, the triangular mesh corresponding to $S$ can be found very efficiently as the solution of a sparse linear system of equations~\cite{vonTycowicz_ea2018}.

\section{A short introduction to geometric statistics in Riemannian manifolds}
\label{app:math_detail}
In the following, whenever we say ``smooth'' we mean ``infinitely often differentiable''.

A (smooth) manifold $M$ of dimension $n$ is a space that \textit{locally} looks like $n$-dimensional Euclidean space. In particular, at each point $p \in M$ the tangent space $T_pM$ is an $n$-dimensional vector space; but in contrast to Euclidean space, the tangent spaces cannot simply be identified with each other (think of the 2-dimensional sphere in $\mathbb{R}^3$ whose tangent spaces are rotated against each other). Thus, keeping track of the base point $p$ is important. For more on this including the exact definitions and examples we refer to \textcolor{black}{Ref.}~\cite[Ch. 0]{doCarmo1992}.

We call $M$ \textit{Riemannian manifold} when it is endowed with a \textit{Riemannian metric}, i.e., a smooth map $ M \ni p \mapsto \langle \cdot, \cdot \rangle_p$ that assigns a scalar product to each tangent space $T_pM$. It allows to measure the length of a smooth curve $\alpha: I \to M$, $I \subset \mathbb{R}$, by defining
$$\text{length}(\alpha) := \int_{I} \langle \alpha'(t), \alpha(t)' \rangle_{\alpha(t)} \textnormal{d} t,$$
where $\alpha' := \frac{\dd}{\dd t} \alpha$ is the tangent vector (field) of $\alpha$. This notion can be used to introduce a distance on $M$.
To this end, let $\Gamma$ be the set of all curves that start in $p \in M$ and end in $q \in M$. The distance between $p$ and $q$ is then given by 
$$d(p,q) := \inf_{\alpha \in \Gamma} \ \text{length}(\alpha).$$
The infimum is necessary because of the possibility that there is no shortest curve connecting $p$ to $q$ (think of a punctured plane in $\mathbb{R}^3$ and $p,q$ such that the missing point is on the straight line connecting them in the plane).

Every Riemannian manifold comes with a notion of a \textit{covariant derivative} along curves. Given a vector field $X$ along a curve $\alpha$ in $M$ (i.e., a map that smoothly assigns to each point on $\alpha$ a vector in the associated tangent space), the covariant derivative allows to differentiate $X$ along $\alpha$; we denote the resulting vector field along $\alpha$ by $\frac{\textnormal{D} X}{\dd t}$.

The covariant derivative induces the notion of ``straightness''. More precisely, \textit{geodesics}---i.e., generalized straight lines---can be defined as curves $\gamma: [0,1] \to M$ whose acceleration vanishes; that is, $\frac{\textnormal{D} \gamma'}{\dd t}$ is the zero vector field.
An important fact is that every point in $M$ has a so-called convex neighborhood $U$. Each pair $p,q \in U$ can be connected by a unique, length-minimizing geodesic $[0,1] \ni t \mapsto \gamma(t;p,q)$ that does not leave $U$. In particular, geodesics are also shortest paths in $U$. (The latter need not be true on a larger scale.) In this work, we always assume that the given data lies in such a convex neighborhood. Then, $\gamma$ is also differentiable with respect to its start and end point. Explicit formulas of these differentials involve the Riemannian curvature tensor $R$ (which intuitively measures local deviation from flat space; see~\cite[Ch. 4]{doCarmo1992}).

The covariant derivative also allows to translate vectors along curves. This is important, e.g., when we want to compare vectors from different tangent spaces. For every pair $p,q \in U$ and $v \in T_pM$ there is a unique vector field $X$ along the geodesic $\gamma$ from $p$ to $q$ that solves the differential equation $\frac{\textnormal{D}X}{\dd t} = 0$ with initial value $X(p) = v$. We call $X(q) \in T_qM$ the \textit{parallel transport} of $v$ to $q$ along $\gamma$ as the vector field does not undergo any intrinsic changes. (Extrinsically---i.e., viewed from outside of the manifold---there might be a change however; this can be seen, e.g., by considering the velocity vector of a \textit{straight-driving} car from the viewpoint of an astronaut in the ISS.)  

Another important map is the \textit{Riemannian exponential}. Let $X \in T_pM$ such that there is a geodesic $[0,1] \ni t \mapsto \gamma(t;p,q)$ in $U$ with $X = \gamma'(0;p,q)$. The exponential map at $p$ is then defined by $\exp_p(X) := q$. Its inverse is the \textit{Riemannian logarithm} $\log_p$. In particular, we find $\log_p(q) = \gamma'(0;p,q)$ and
\begin{equation*} 
    d(p,q) = \| \log_p(q) \| = \sqrt{\langle \log_p(q), \log_p(q) \rangle_p}.
\end{equation*}
Because of this and the fact that it is parallel to the shortest path from $p$ to $q$ (i.e., ``points'' to $q$), $log_p(q)$ can be seen as the ``difference vector'' between $q$ and $p$. 

In the differential coordinate space, the exponential and logarithm map as well as parallel transport can be computed using explicit formulas: Because of the product structure, they can be calculated component-wise for every element of SO(3) and $\Symp(3)$; formulas for these manifolds are derived in \textcolor{black}{Refs.}~\cite{EdelmanAriasSmith1998, Arsigny_ea2007}, respectively.
\bigskip

Amongst others, the tools gathered so far allow to generalize the notion of mean and linear regression to a Riemannian manifold $M$ in the form of the Fr\'{e}chet mean and geodesic regression, respectively. 
\begin{figure*}[ht]
\centering
    \centering
    \includegraphics[width=.85\textwidth]{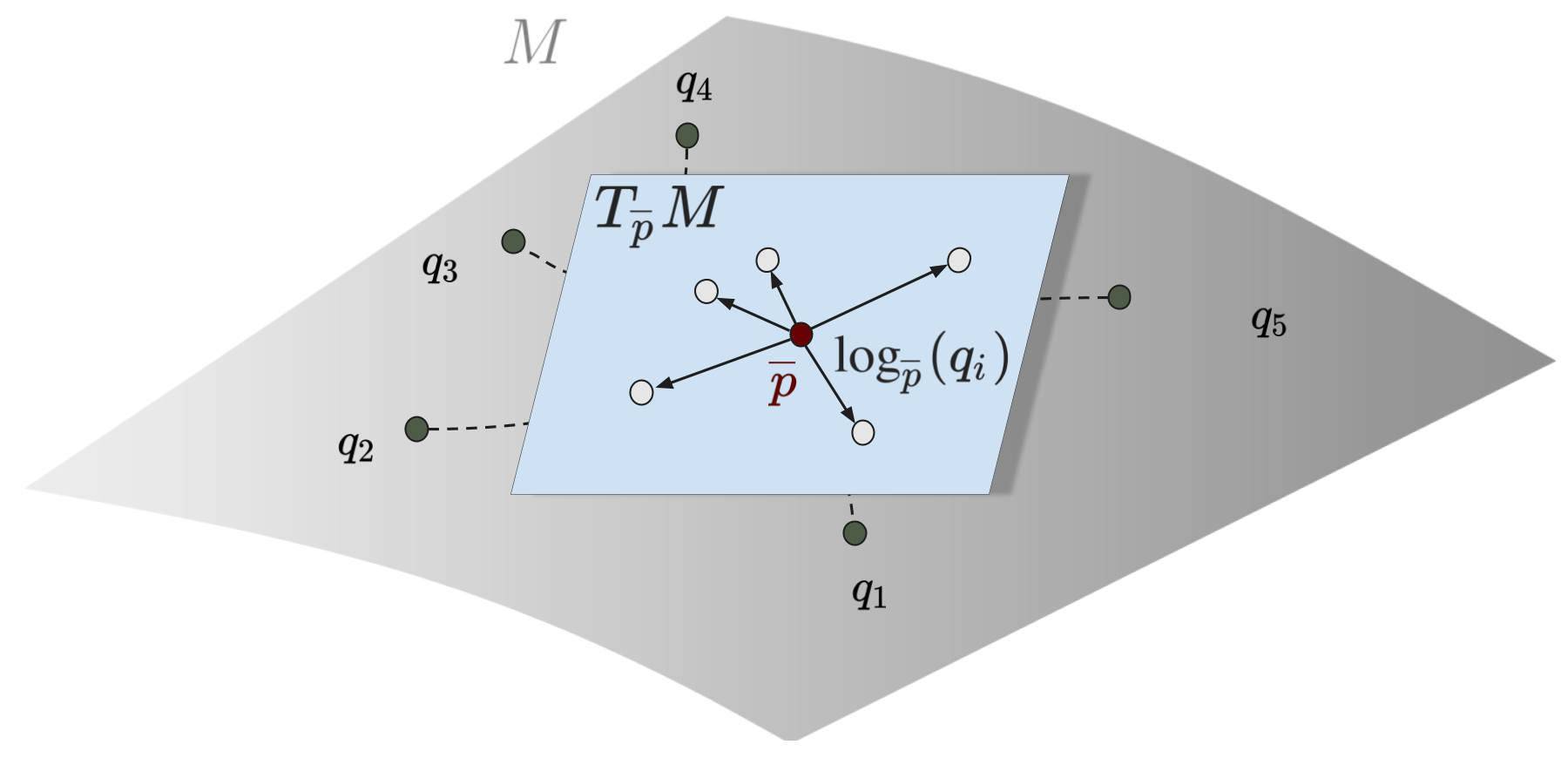}
    \caption{Fr\'{e}chet mean of $q_1,\dots,q_5$ in a Riemannian manifold $M$. After mapping the points into the tangent space $T_{\overline{p}}M$ at the Fr\'{e}chet mean $\overline{p}$ with $\log_{\overline{p}}$, the resulting tangent vectors $\log_{\overline{p}}(q_i)$, $i=1,\dots,5$ sum to zero. The broken lines are the geodesics from the mean to the data points.} \label{fig:frechet_mean}
    \Description[Fr\'{e}chet mean]{Visualization of the barycentric property of the Fr\'{e}chet mean in an arbitrary Riemannian manifold}
\end{figure*}
Given data points $q_1,\dots,q_n$ in a convex neighborhood $U \subseteq M$, their \textit{Fr\'{e}chet mean} $\overline{p} \in U$ is defined implicitly by 
$$\sum^n_{i=1} \log_{\overline{p}}(q_i) = 0.$$ 
That is, we look for a point $\overline{p} \in U$ in whose tangent space the usual optimality condition of the multivariate mean holds---namely, that the sum of difference vectors to the data points vanishes. This idea is visualized in Fig.~\ref{fig:frechet_mean}. Importantly, Fr\'{e}chet means can be computed efficiently using an iterative scheme. (For more on Fr\'{e}chet means including a discussion of their existence, we refer the reader to \textcolor{black}{Ref.~}\cite[Sect.~2.2]{Fletcher2020}.)

\begin{figure*}[ht]
\centering
    \centering
    \includegraphics[width=.85\textwidth, trim={0 0 0 1.3cm},clip]{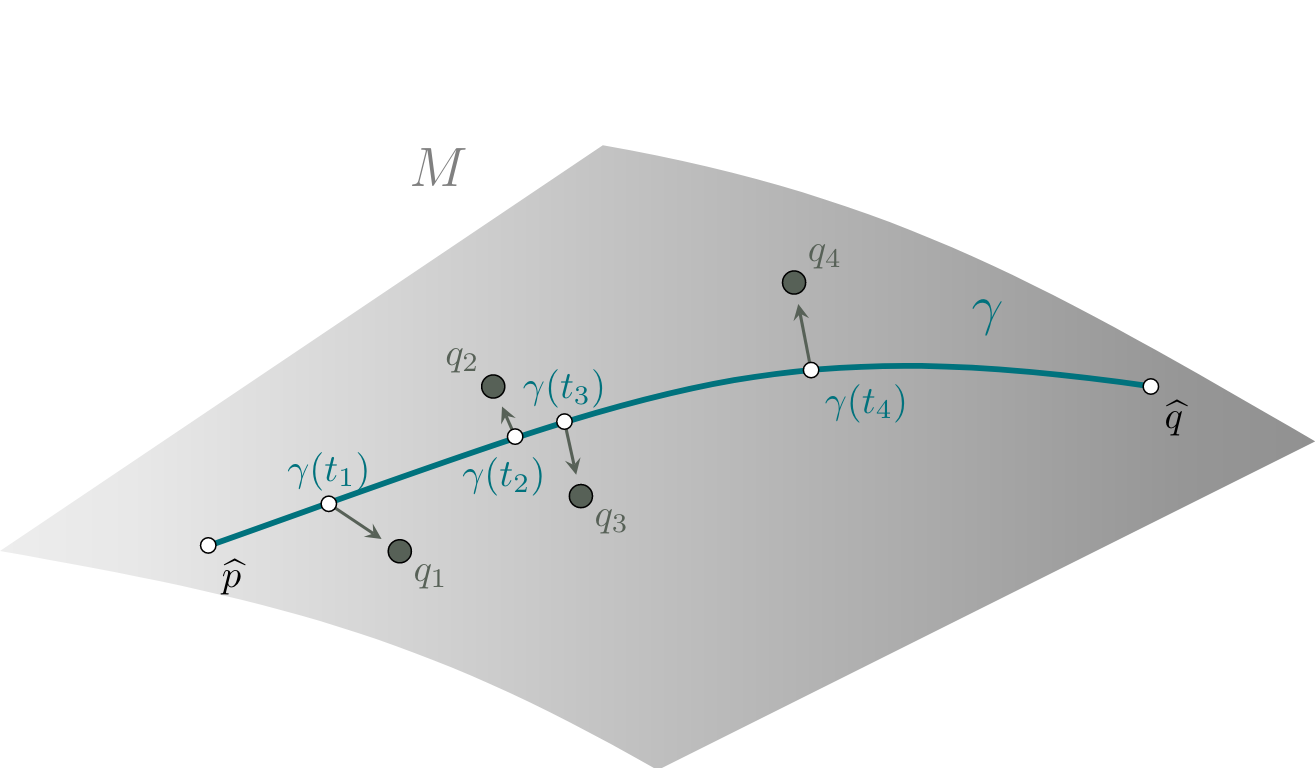}
    \caption{Geodesic regression in (a part of) a Riemannian manifold $M$ w.r.t.\ data $(q_1, t_1),\dots, (q_4, t_4)$. The geodesic $\gamma$ is the least-squares estimator with start and end point $\widehat{p}$ and  $\widehat{q}$, respectively. The gray arrows are the vectors $\log_{\gamma(t_i)}(q_i) \in T_{\gamma(t_i)}M$.} \label{fig:geodesic_regression_synthetic}
    \Description[Geodesic regression]{Visualization of the minimizing property of the optimal geodesic in an arbitrary Riemannian manifold}
\end{figure*}

Given $n$ data points $(q_i, t_i) \in U \times \mathbb{R}$ together with parameter values, \textit{geodesic regression}~\cite{Fletcher2013} assumes that the $q_i$ are realizations of an $M$-valued random variable $Q$ according to the model
\begin{equation} \label{eq:model}
    Q(t) := \exp_{\gamma(t;p,q)}(\epsilon),
\end{equation}
where $\epsilon$ is tangent-vector-valued noise and $p,q \in U$. The goal is to find the geodesic $\gamma$ in (\ref{eq:model}). A good approximation in many cases is the \textit{least-squares estimator}. Since geodesics in $U$ can be parametrized by their start and endpoint, it is the minimizer $(\widehat{p}, \widehat{q}) \in U \times U$ of the \textit{sum-of-squared error}
$$\mathcal{E}(p,q) := \sum_{i=1}^n d\big(\gamma(t_i;p,q), q_i \big)^2.$$
In the spaces that are relevant to us, $(\widehat{p}, \widehat{q})$ can be found efficiently using optimization methods. 
In particular, in the space of differential coordinates (which, as a product of symmetric spaces, is also a symmetric space~\cite{Helgason2001}) an explicit formula for the intrinsic gradient of $\mathcal{E}$ is available~\cite{BergmanGousenbourger2018,Hanik_ea2020}, so that we can use Riemannian gradient descent~\cite{AbsilMahonySepulchre2007} to compute the minimizer.
A visualization of geodesic regression with respect to\ 4 data points is depicted in Fig.~\ref{fig:geodesic_regression_synthetic}.

\begin{acks}
    M. Hanik is funded by Deutsche Forschungsgemeinschaft (DFG, German Research Foundation) under Germany's Excellence Strategy through MATH+, the Berlin Mathematics Research Center, under Grant No.: EXC-2046/1 -- project ID: 390685689. This work was also supported by the Bundesministerium f\"ur Bildung und Forschung (BMBF) through BIFOLD, the Berlin Institute for the Foundations of Learning and Data under Grant Nos.:~ 1IS18025A and 01IS18037A. 
    
    Last but not least, we thank the anonymous reviewers for their insightful comments and helpful suggestions.
\end{acks}

\bibliographystyle{ACM-Reference-Format}
\bibliography{mybibfile}

\newpage

\begin{table}[H]
    \caption{{\bf Spherical sundials used in the study}}
    \centering 
    \begin{longtable}{ |c c|c c|c|c|c| } 
        \hline
        Object ID & Dialface ID & Latitude & Longitude & Site & Dating & \parbox{3cm}{\vspace*{1mm}\centering Type \\ (Edition Topoi)\vspace*{1mm}}\\
        \hline\hline
            &  & & & \textbf{Italy} & &\\
        \hline\hline
        574 & 623 & 43.3155 & 13.4082 & Helvia Recina &      N/A        & \parbox{3cm}{\centering sphere, spherical \\ }\\
        \hline
        17  & 17  & 42.0913 & 12.5231 & Riano  &      N/A               & \parbox{4cm}{\vspace*{1mm}\centering sphere, spherical-cut dials---central gnomon point\vspace*{1mm}} \\ 
        \hline
        36  & 35  & 41.8034 & 12.6890 & \parbox{4cm}{\vspace*{1mm}\centering Villa Tuscolana/Villa Rufinella\footnotemark[5] \\ (near Rome)\vspace*{1mm}} & N/A  & \parbox{4cm}{\vspace*{1mm}\centering sphere, spherical-cut dials---central gnomon point\vspace*{1mm}}\\ 
        \hline
        65  & 62  & 41.7561 & 12.2927 & \parbox{4cm}{\vspace*{1mm}\centering Ostia Antica \\ (near Rome)\vspace*{1mm}}  &   N/A    & \parbox{4cm}{\vspace*{1mm}\centering sphere, spherical-cut/ \\ quarter dis-shaped\vspace*{1mm}} \\
        \hline
        18  & 18  & 41.6700 & 12.6900 & Lanuvio  &  N/A  & \parbox{4cm}{\vspace*{1mm}\centering sphere, spherical-cut dials---central gnomon point\vspace*{1mm}}\\ 
        \hline
        21  & 21  & 40.7503 & 14.4871 & Pompei &  \parbox{1cm}{\centering before \\ 79 CE}    & \parbox{4cm}{\vspace*{1mm}\centering sphere, spherical-cut dials---central gnomon point\vspace*{1mm}}\\ 
        \hline
        23  & 23  & 40.7503 & 14.4871 & Pompei  &   \parbox{1cm}{\vspace*{1mm}\centering before \\ 79 CE\vspace*{1mm}}  & \parbox{4cm}{\vspace*{1mm}\centering sphere, spherical-cut dials---central gnomon point\vspace*{1mm}}\\
        \hline
        174 & 173 & 40.7503 & 14.4871 & Pompei  &  N/A  & \parbox{4cm}{\vspace*{1mm}\centering sphere, spherical-hemispherical\\
        ---central gnomon point\vspace*{1mm}}\\ 
        \hline
        29  & 29  & 40.7503 & 14.4871 & Pompei  &  \parbox{1cm}{\vspace*{1mm}\centering before \\ 79 CE\vspace*{1mm}}   & \parbox{4cm}{\vspace*{1mm}\centering sphere, spherical-cut dials---central gnomon point\vspace*{1mm}} \\
        \hline
        519 & 559 & 40.7030 & 14.4988 & Stabiae &  N/A   & \parbox{4cm}{\centering sphere, spherical \\ }\\
        \hline
        40  & 39  & N/A & N/A        & N/A\footnotemark[6]     &  N/A   & \parbox{4cm}{\vspace*{1mm}\centering sphere, spherical-cut dials---central gnomon point\vspace*{1mm}}\\
        \hline\hline
             & & & & \textbf{Greece} & &\\
        \hline\hline
        76  & 73  & 37.3900 & 25.2600 & Delos& N/A & \parbox{4cm}{\vspace*{1mm}\centering sphere, spherical-hemispherical---central \\ gnomon point\vspace*{1mm}} \\ 
        \hline
        77  & 74  & 37.3900 & 25.2600 & Delos&  N/A & \parbox{4cm}{\vspace*{1mm}\centering sphere, spherical-transposed \\ hemispherical\vspace*{1mm}} \\
        \hline
        546 & 583 & 36.0917 & 28.0881 & \parbox{4cm}{ \vspace*{1mm}\centering Lindos on \\ Rhodes\footnotemark[7]\vspace*{1mm}}& hellenistic & \parbox{4cm}{\centering sphere, spherical \\ }\\[1.5ex] 
        \hline
    \end{longtable}\label{table2}
\end{table}

\footnotetext[5]{This site is not 100\% certain; the only reference comes from a dissertation written in 1764.}
\footnotetext[6]{Although in the data base the site is stated to be Vatican City, we did not find evidence for this. Therefore, we consider the site to be uncertain.}
\footnotetext[7]{The longitude and latitude of this sundial are wrong in the database. The references stated there confirm that the sundial is from Lindos. We have used the correct values, specified in the table.}

\end{document}